\documentclass{article}

\usepackage{arxiv}
\usepackage{multirow}
\usepackage[utf8]{inputenc} 
\usepackage[T1]{fontenc}    
\usepackage{hyperref}       
\usepackage{url}            
\usepackage{booktabs}       
\usepackage{amsfonts}       
\usepackage{nicefrac}       
\usepackage{microtype}      
\usepackage{lipsum}
\usepackage{graphicx}
\usepackage{algorithm}
\usepackage{algorithmic}
\usepackage{amsmath}
\usepackage{subcaption} 
\usepackage{adjustbox} 
\usepackage{amssymb,amsthm}
\usepackage[mathscr]{euscript}
\usepackage{appendix}
\graphicspath{ {./images/} }

\title{Conformal P-Value in Multiple-Choice Question Answering Tasks with Provable Risk Control}

\author{
 Yuanchang Ye \\
  School of Data Sciences\\
  Zhejiang University of Finance \& Economics\\
  HangZhou, China \\
  \texttt{yuanchang0213@zufe.edu.cn} \\
}

\begin{document}
\maketitle
\begin{abstract}
This study introduces a significance testing-enhanced conformal prediction (CP) framework to improve trustworthiness of large language models (LLMs) in multiple-choice question answering (MCQA). While LLMs have been increasingly deployed in disciplinary QA scenarios, hallucination and nonfactual generation substantially compromise response reliability. Although CP provides statistically rigorous marginal coverage guarantees for prediction sets, and significance testing offers established statistical rigor, their synergistic integration remains unexplored. 
To mitigate hallucination and factual inaccuracies, our framework integrates $p$-value computation with conformity scoring through self-consistency resampling of MCQA responses. This approach calculates option frequencies to address LLMs' black-box nature, subsequently constructing prediction sets via null hypothesis testing ($\mathcal{H}_0$) with empirically derived $p$-values. 
Evaluations on MMLU and MMLU-Pro benchmarks using off-the-shelf LLMs demonstrate: (1) The enhanced CP achieves user-specified empirical miscoverage rates; (2) Test-set average prediction set size (APSS) decreases monotonically with increasing risk levels ($\alpha$), validating APSS as an effective uncertainty metric. This work establishes a principled statistical framework for trustworthy LLM deployment in high-stakes QA applications.
\end{abstract}


\section{Introduction}
The rapid advancement of large language models (LLMs) has enabled their widespread deployment in domains including AI-powered customer service, content generation, and knowledge QA~\cite{bi-etal-2025-llava,bi2025cot,chen2025fedbip,zhang2023spot}. Nevertheless, despite continuous improvements in natural language generation capabilities, studies indicate that LLMs persistently exhibit factual hallucinations, confidently producing erroneous or non-factual text. Such outputs often demonstrate logical coherence, authoritative presentation, and high persuasiveness while containing factual deviations or fictitious content. Direct utilization of hallucinated content risks erroneous decision-making, user misinformation, and ultimate compromise of trustworthiness in high-stakes applications~\cite{bi2025prism,chen2025does,rong2025backdoor}. Consequently, developing efficient, automated hallucination detection frameworks constitutes a critical challenge for ensuring LLM reliability and application safety~\cite{madhusudhan2024llms}.

Uncertainty quantification (UQ) is essential for assessing model reliability, risk management, and hallucination identification~\cite{wang2025ascd,wang2025word,wang2025coin}. Since LLMs primarily provide text-based interfaces, prevailing confidence assessment methods include calibration-based techniques, verbalized uncertainty expressions, and heuristic approaches. However, these heuristic methods cannot offer task-specific performance guarantees, limiting their reliability. While conformal prediction (CP) offers model-agnostic, statistically rigorous uncertainty estimation, its application to natural language generation (NLG) remains challenging. To address this, we adapt statistical significance testing methodologies, formulating rigorous mathematical constructs that ensure performance guarantees while maintaining statistical robustness.

Our Significance Testing-based Conformal Prediction (ST-CP) approach differs fundamentally from conventional CP methods~\cite{angelopoulos2021gentle} by delivering statistical coverage guarantees under minimal assumptions while achieving superior computational efficiency. In this work, we implement ST-CP for Multiple Choice Question Answering (MCQA) tasks~\cite{ye2024benchmarking}. The data is partitioned into calibration and test sets based on a predefined ratio. Following multiple resampling iterations of question responses, we compute p-values using statistical significance tests. These p-values are then evaluated against a predetermined significance level ($\alpha$) to determine null hypothesis rejection—thereby governing answer choice inclusion in the prediction set.

We evaluate our approach using MCQA datasets MMLU and MMLU\_PRO, testing four LLM variants: Qwen2.5\-3B-Instruct, Llama\-3.2\-3B\-Instruct, Meta-Llama-3-8B-Instruct, and Vicuna\-7B\-v1.5. To satisfy CP's exchangeability condition, experiments are performed within each subject-specific dataset. Extensive empirical results demonstrate precise control of the miscoverage rate across user-specified risk levels. For example, on the MMLU\_PRO biology benchmark with elevated risk tolerance ($\alpha = 0.5$), Vicuna\-7B\-v1.5 maintains an average empirical error rate below $\alpha = 0.5$. Notably, the method consistently constrains the empirical error rate below predefined risk levels irrespective of calibration-test split ratios. This confirms that even with limited calibration data, the framework preserves coverage guarantees and exhibits strong robustness. Moreover, increasing the risk parameter $\alpha$ progressively reduces the average prediction set size. This behavior demonstrates our method's capacity to produce statistically rigorous, compact prediction sets that effectively reduce hallucination propensity in LLM outputs.
\section{Related Work}
\textbf{Hallucinations in LLMs:} Within natural language processing, hallucination denotes model-generated content that appears coherent but diverges from source inputs (e.g., prompts, context) or contradicts verifiable world knowledge—a concept adapted from psychological studies on perceptual distortions. In task-specific settings like multiple-choice question answering (MCQA), LLM hallucinations manifest prominently as two core types: intrinsic hallucinations (answers conflicting directly with question context or options) and extrinsic hallucinations (answers relying on unverifiable external knowledge without explicit context contradiction). Given LLMs’ core objective of accurate instruction adherence (e.g., selecting correct answers), MCQA hallucinations further categorize into: factual hallucinations (deviations from objective/question-specific verifiable facts) and faithfulness hallucinations (failures in semantic alignment with question constraints, including option misinterpretation or logical incoherence).

Detection methodologies primarily comprise: 1) External model-based evaluation, employing high-precision discriminators (e.g., superior LLMs) to score responses—limited by annotation quality/availability; 2) Rule-based detection, targeting dimension-specific hallucination errors (e.g., factual/entity inaccuracies) using established benchmarks like LAMA, TruthfulQA, and FaithDial for quantification.

Nevertheless, LLMs confront persistent challenges in hallucination mitigation: Techniques like Contrastive Decoding address cognitive biases via candidate comparison or preference modeling but exhibit limited efficacy in MCQA due to distractor interference and contrast instability; Post-hoc correction methods (e.g., iterative prompting) incur prohibitive computational costs and lack cross-task generalizability. Fundamentally, LLMs’ parametric knowledge representation and generative mechanisms perpetuate vulnerabilities including ambiguous knowledge boundaries, brittle reasoning chains, and hypersensitivity to noise. Consequently, hallucination risks endure in precision-demanding MCQA tasks. This framework offers a systematic approach for analyzing and enhancing content reliability in LLMs, particularly for rigorous QA applications.

\textbf{Integrated SCP-Significance Testing Framework:} Split Conformal Prediction (SCP) and Significance Testing constitute synergistic~\cite{wang-etal-2025-sconu}, statistically rigorous paradigms for uncertainty quantification in LLM hallucination research. SCP leverages exchangeable calibration data to construct prediction sets with guaranteed coverage probability ($\ge 1-\alpha$) for black-box model outputs. Its model-agnostic and distribution-free properties require only data exchangeability. For hallucination detection, SCP dynamically filters generated candidates (e.g., high-confidence erroneous entities) via confidence thresholds or truncates low-likelihood outputs using stopping rules~\cite{wang2024conu,wang2025sample}.

Significance Testing identifies hallucinations through hypothesis tests~\cite{gui2024conformal,wang2024conformalized} (e.g., p-value computation for generation probabilities), detecting outputs substantially deviating from verifiable knowledge/context. Key applications include: 1) False Discovery Rate (FDR) control via multiple testing correction (e.g., Benjamini\-Hochberg procedure for hallucinated entity identification); 2) Multi-source consistency evaluation using combined tests (e.g., Fisher’s method).

The integrated framework operates bidirectionally: SCP delimits candidate spaces to streamline Significance Testing, while significance outcomes dynamically optimize SCP’s confidence thresholds—enhancing coverage reliability by controlling hallucination omission rates. Despite challenges in non-exchangeable data adaptation, multi-step error propagation, and computational overhead, SCP\-Significance Testing furnishes a theoretically grounded and practically viable framework for hallucination quantification, risk mitigation, and reliability assessment in LLMs, capitalizing on rigorous statistical guarantees and model-agnosticism.
\section{Method}
Our method primarily addresses two challenges: (1) How to construct prediction sets with marginal guarantees on test data using significance testing and Conformal Prediction (CP), and (2) How to demonstrate that our approach satisfies statistical significance requirements. In this section, we first introduce essential notation definitions. Subsequently, we present the fundamental Conformal Prediction framework, through which we construct p-values for significance testing. Finally, we provide theoretical proofs for achieving marginal guarantees.

\begin{algorithm}[!h]
    \caption{Genarate Prediction Set}
    \label{alg:AOS}
    \renewcommand{\algorithmicrequire}{\textbf{Input:}}
    \renewcommand{\algorithmicensure}{\textbf{Output:}}
    
    \begin{algorithmic}[1]
        \REQUIRE $(x_{i},y_{i}^{*})^{n}_{i=1}$,$(x_{test},y)$,$\alpha$  
        \ENSURE prediction set $\mathcal{C}_{\alpha}(x_{test})$    
        
        \STATE  // Construct calibration data
        \STATE Initialize $\mathcal{D}_\mathrm{cal}=\{(x_{i},y_{i}^{*})\}_{i=1}^{n}$
        \FOR{each $x_i \in \mathcal{D}$}
            \STATE $s_i \leftarrow 1-\hat{f}(x_i,y^*)$
        \ENDFOR 

        \STATE sort $\{s_i\}_{i=1}^N$
        \FOR{each $y_k^{test} \in \mathcal{Y}$}
            \STATE calculate $S(x_{test},y_k^{test})$
            \IF {$\mathbb{P}\left(\frac{\sum_{i=1}^n \mathbf{1}\{s_i > s_{\text{test}}\} + 1}{n+1} \leq \alpha\right)  \leq \alpha$}
                \STATE continue
            \ELSE
                \STATE $\mathcal{C}_{\alpha}(x_{test}) \leftarrow S(x_{test},y_k^{test})$
            \ENDIF
        \ENDFOR
        
        \RETURN Prediction Set $\mathcal{C}_{\alpha}(x_{test})$
    \end{algorithmic}
\end{algorithm}

\subsection{Preliminary}
\label{subsec:prelim}
We formulate the task as a multiple-choice question answering (MCQA) problem with $K$ distinct options. Following conventional CP methodology, let $\mathcal{D}_\mathrm{cal}=\{(x_{i},y_{i}^{*})\}_{i=1}^{n}$ denote the calibration dataset containing ground-truth labels, where $n$ represents the total number of calibration
examples and $y_i^*$ denotes the true label for the $i$-th instance. We consider a test input-output pair
$(x_{n+1},y_{n+1})$ for prediction. The significance level $\alpha\in(0,1)$ serves dual purposes: determining
the Typel error rate in hypothesis testing and specifying the desired coverage rate in CP. We treat the Large Language Model (LLM) as a black-box predictor $\hat{f}:\mathcal{X}\to\mathcal{Y}$ mapping input features to output labels, remaining agnostic to its internal mechanisms. A critical component in the CP
framework is the non-conformity score $S:\mathcal{X}\times\mathcal{Y}\to\mathbb{R}$, which heuristically measures the
compatibility between inputs and candidate labels. For classification tasks, we define $\mathcal{S}(x,y)=$
$1-\hat{f}(y|x)$ where $\hat{f}(y|x)$ represents the model's estimated probability for class $y$,with $S_i=$
$\mathcal{S}(x_i,y_i)$ denoting the non-conformity score for the $i$-th calibration example.

\subsection{Conformal Prediction}
\label{subsec:cp}
For each calibration example $(x_{i},y_{i}^{*})$, we perform $P$ independent samplings from the LLM to obtain response sets $\{\hat{y}_{1}^{(i)},\hat{y}_{2}^{(i)},\hat{y}_{3}^{(i)},\ldots,\hat{y}_{K}^{(i)}\}$. We calculate the empirical frequency of ground-truth label occurrences as $\hat{f}(y|x)$, subsequently computing the non-conformity score $S(x_{i},y^{*})$. For MCQA problems with multiple valid answers, $\hat{f}(y|x)$ corresponds to the aggregate frequency of all correct options. The procedure for generating conformal prediction sets for unseen test instances $x_{{test}}$ comprises three stages:

\begin{enumerate}
    \item Compute non-conformity scores $\{s_{1}, \ldots, s_{n}\}$ for calibration data where $s_{i}=S(x_{i}, y_{i}^{*})$
    \item Determine the conformal $\alpha$-quantile $\tau=Q_{1-\alpha}(\{s_{i}\}_{i=1}^{n})$ from the empirical score distribution. Following standard CP practice, we calculate $\tau$ as the $[(1-\alpha)(n+1)]$-th smallest value in the sorted calibration scores
    \item Construct the prediction set: 
\begin{equation}
    C_{\alpha}(x_{\text{test}})=\left\{y: S(x_{\text{test}}, y) \leq \text{Q}\left(\{s_{i}\}_{i=1}^{n}, \frac{[(n+1)(1-\alpha)]}{n}\right)\right\}
\end{equation}
This guarantees: 
\begin{equation}
\mathbb{P}\left(y_{\text{test}}^{*} \in C_{\alpha}(x_{\text{test}})\right) \geq 1-\alpha
\end{equation}
\end{enumerate}

Steps 1-2 constitute the calibration phase, while Step 3 generates the final prediction set. Intuitively, the set includes all labels whose compatibility with the test input exceeds the $\alpha$-quantile threshold established by the calibration distribution.

\subsection{Significance Testing}
\label{subsec:sigtest}

Following fundamental principles of statistical significance testing, we establish a p-value for comparison with our predefined significance level $\alpha$. By reformulating Equation (1) using the empirical cumulative distribution function (ECDF), we derive:
\begin{equation}
    \mathcal{C}_\alpha(x_{\text{test}}) = \left\{ y: \frac{1}{n}\sum_{i=1}^n \mathbf{1}\{s_i \leq S(x_i,y)\} \leq \frac{\lceil (n+1)(1-\alpha) \rceil}{n} \right\}
\end{equation}

where $\mathbf{1}\{\cdot\}$ denotes the indicator function. This formulation implies that a label $y$ is included if the proportion of calibration scores $\{s_i\}$ smaller than its non-conformity score $S(x_{\text{test}},y)$ does not exceed the adjusted threshold $\frac{\lceil (n+1)(1-\alpha) \rceil}{n}$. Through algebraic manipulation of Equation (3), we obtain the equivalent condition:

\begin{equation}
 \frac{\sum_{i=1}^n \mathbf{1}\{s_i > S(x_{\text{test}},y)\} + 1}{n+1} > \alpha
\end{equation}

This equivalence establishes our hypothesis testing framework with:

\begin{itemize}
    \item Null hypothesis $\mathcal{H}_0$: $y$ is the true label of $x_{\text{test}}$
    \item Alternative hypothesis $\mathcal{H}_1$: $y$ is not the true label
\end{itemize}

We define the p-value statistic as:

\begin{equation}
p(y) = \frac{\sum_{i=1}^n \mathbf{1}\{s_i > S(x_{\text{test}},y)\} + 1}{n+1}
\end{equation}

For the ground-truth label $y_{\text{test}}^*$, this simplifies to:
\begin{equation}
 p(y_{\text{test}}^*) = \frac{\sum_{i=1}^n \mathbf{1}\{s_i > s_{\text{test}}\} + 1}{n+1}, \quad s_{\text{test}} = S(x_{\text{test}}, y_{\text{test}}^*)
\end{equation}

The decision rule follows standard significance testing principles: we reject $\mathcal{H}_0$ when $p(y) \leq \alpha$, thereby excluding $y$ from the prediction set. The coverage guarantee emerges from the probabilistic bound:

\begin{equation}
\frac{\sum_{i=1}^{N}\mathbf{1}\{s_{i}>s_{N+1}\}+1}{N+1}\leq\alpha
\end{equation}

The equivalence between Equations (7) and (2) demonstrates that our method satisfies the marginal coverage guarantee through its connection to conformal prediction theory. This establishes the statistical validity of our framework under standard exchangeability assumptions.
\section{Experiments}
\subsection{Experimental Settings}
\subsubsection{Datasets}
Our experiments utilize MMLU and its augmented variant MMLU-Pro as primary multiple-choice question answering (MCQA) benchmarks. MMLU comprises 15,908 questions spanning 57 subjects (including STEM, humanities, social sciences), partitioned into few-shot development (285 questions), validation (1,540 questions), and test sets (14,079 questions) for evaluating zero-shot/few-shot knowledge proficiency. MMLU-Pro serves as an advanced-difficulty benchmark containing 12,032 questions, with 5,222 questions curated from three high-quality sources (STEM websites/TheoremQA/SciBench) to augment coverage in nine core disciplines—notably physics (+888), chemistry (+954), and engineering (+902). The benchmark's innovative ten-option design (versus MMLU's four options) significantly elevates discriminative difficulty. Collectively, these benchmarks provide rigorous evaluation platforms for hallucination detection and knowledge boundary evaluation through their disciplinary diversity, advanced reasoning demands, and stringent partitioning protocols.

\begin{table}[t]
\centering
\caption{Average Error Rate for Different Datasets and Models under split-ratio=0.5}
\adjustbox{max width=\linewidth}{
\begin{tabular}{ccccccccccc} 
\hline
\multirow{2}{*}{Dataset}                       & \multirow{2}{*}{Model}   & \multicolumn{9}{c}{Alpha Rate}                                                  \\ 

                                               &                          & 0.1    & 0.2    & 0.3    & 0.4    & 0.5    & 0.6    & 0.7    & 0.8    & 0.9     \\ 
\hline
\multirow{7}{*}{MMLU}                          & Llama-3.1-8B-Instruct    & 0.088  & 0.187  & 0.288  & 0.389  & 0.488  & 0.584  & 0.684  & 0.785  & 0.877   \\
                                               & Llama-3.2-3B-Instruct    & 0.090  & 0.187  & 0.284  & 0.388  & 0.487  & 0.582  & 0.681  & 0.782  & 0.877   \\
                                               & Meta-Llama-3-8B-Instruct & 0.091  & 0.192  & 0.288  & 0.392  & 0.490  & 0.585  & 0.686  & 0.785  & 0.876   \\
                                               & Qwen2.5-3B-Instruct      & 0.091  & 0.190  & 0.290  & 0.392  & 0.488  & 0.584  & 0.683  & 0.783  & 0.879   \\
                                               & Qwen2.5-7B-Instruct      & 0.090  & 0.189  & 0.289  & 0.389  & 0.486  & 0.582  & 0.683  & 0.785  & 0.877   \\
                                               & openchat\_3.5            & 0.088  & 0.186  & 0.284  & 0.386  & 0.483  & 0.579  & 0.679  & 0.782  & 0.876   \\
                                               & vicuna-7b-v1.5           & 0.091  & 0.191  & 0.289  & 0.391  & 0.490  & 0.582  & 0.684  & 0.782  & 0.875   \\ 
\hline
\multicolumn{1}{l}{\multirow{7}{*}{MMLU-PRO}} & Llama-3.1-8B-Instruct    & 0.099  & 0.197  & 0.296  & 0.395  & 0.494  & 0.596  & 0.693  & 0.793  & 0.893   \\
\multicolumn{1}{l}{}                          & Llama-3.2-3B-Instruct    & 0.098  & 0.197  & 0.297  & 0.397  & 0.496  & 0.597  & 0.697  & 0.796  & 0.893   \\
\multicolumn{1}{l}{}                          & Meta-Llama-3-8B-Instruct & 0.097  & 0.197  & 0.297  & 0.397  & 0.498  & 0.597  & 0.697  & 0.796  & 0.895   \\
\multicolumn{1}{l}{}                          & Qwen2.5-3B-Instruct      & 0.097  & 0.196  & 0.296  & 0.397  & 0.496  & 0.595  & 0.695  & 0.794  & 0.894   \\
\multicolumn{1}{l}{}                          & Qwen2.5-7B-Instruct      & 0.097  & 0.196  & 0.295  & 0.395  & 0.495  & 0.595  & 0.696  & 0.794  & 0.894   \\
\multicolumn{1}{l}{}                          & openchat\_3.5            & 0.099  & 0.199  & 0.298  & 0.396  & 0.497  & 0.596  & 0.695  & 0.796  & 0.894   \\
\multicolumn{1}{l}{}                          & vicuna-7b-v1.5           & 0.098  & 0.199  & 0.298  & 0.398  & 0.497  & 0.596  & 0.696  & 0.795  & 0.893   \\
\hline
\end{tabular}
}
\end{table}

\subsubsection{LLMs}
We conduct comprehensive evaluations using four state-of-the-art language models: Qwen2.5-3B-Instruct, Llama-3-3B-Instruct, Meta-Llama-3-8B-Instruct, and Vicuna-7B-v1.5, with subsequent performance visualization. Architecture-specific comparative analysis is performed across parameter variants. Key configurations include:
\begin{itemize}
    \item Qwen2.5-3B-Instruct: Features Grouped-Query Attention (GQA) and SwiGLU activations for efficient decoding. Implements a three-stage alignment pipeline (multilingual pretraining → supervised fine-tuning → RLHF optimization), optimized for 128K-context complex instruction reasoning.
    \item Llama-3-3B-Instruct: Integrates KV caching and RMSNorm standardization. Trained via two-phase paradigm (pretraining + PPO-RLHF alignment) for efficient 8K-context multilingual processing.
    \item Meta-Llama-3-8B-Instruct: Enhances representational capacity through expanded hidden dimensions (6,656). Combines DPO-RLHF optimization to achieve state-of-the-art reasoning performance in the 8B parameter class.
    \item Vicuna-7B-v1.5: Fine-tuned exclusively on conversational data (125K multi-turn dialogues), exhibiting strong coherence in 4K-context interactions.
\end{itemize}
\subsubsection{Implementation Details}
Marginal coverage guarantees for MCQA prediction sets are implemented through integrated significance testing and Split Conformal Prediction (SCP). The methodological pipeline comprises:(1) Response Sampling: 20 independent generations per question (temperature=1.0, top-p=0.9), with output length constrained to single-token responses for multiple-choice uncertainty quantification.(2) Data Partitioning: Subject-specific datasets are divided into calibration and test sets using predetermined ratios via SCP, with option probabilities derived from model outputs.(3) Statistical Testing: Nonconformity scores are computed from calibration data. Significance testing determines p-values for test-set options, with null hypothesis rejection (option exclusion criterion) occurring when p-values fall below threshold $\alpha$.(4) Evaluation Metrics: Empirical error rates are calculated against ground-truth labels. Prediction set reliability is quantified through 100 bootstrap resampling iterations generating box plots, while prediction set size distributions across significance levels provide complementary assessment of marginal coverage guarantees.

\subsection{Empirical Error Rate}

This section rigorously validates the calibrated prediction sets constructed via Equation (3), confirming their capability to reliably achieve target coverage levels across user-specified miscoverage rates. We further examine the framework's practical utility through uncertainty metric-guided selective prediction.

Empirical verification of nominal coverage levels proceeds by partitioning each dataset into calibration and test sets at a 1:1 ratio. Conformal uncertainty thresholds are derived from calibration data according to prespecified miscoverage rates. Subsequent coverage evaluation on test data yields the MMLU and MMLU-Pro results depicted in Figures 2-4.

First, we set the split-ratio of the data to 0.5. We measured the accuracy of different subjects under each dataset through 100 samplings. Finally, we aggregated and averaged the accuracy of all subjects, and extended the scope to models with more parameters. We got the following results: Under different datasets and models, the error rate of the prediction set is lower than our set significance level. This shows that under our relatively standard split-ratio, the obtained results meet the marginal guarantee. Moreover, all results are not significantly lower than our set significance level, but are lower than the set value within a small range. This ensures that it does not exceed the upper bound nor fall below the lower bound of this method.

\begin{figure}[!t]
\centering 

\begin{adjustbox}{width=\textwidth}
\begin{tabular}{cccc}
\subfloat[]{\includegraphics[width=5\textwidth]{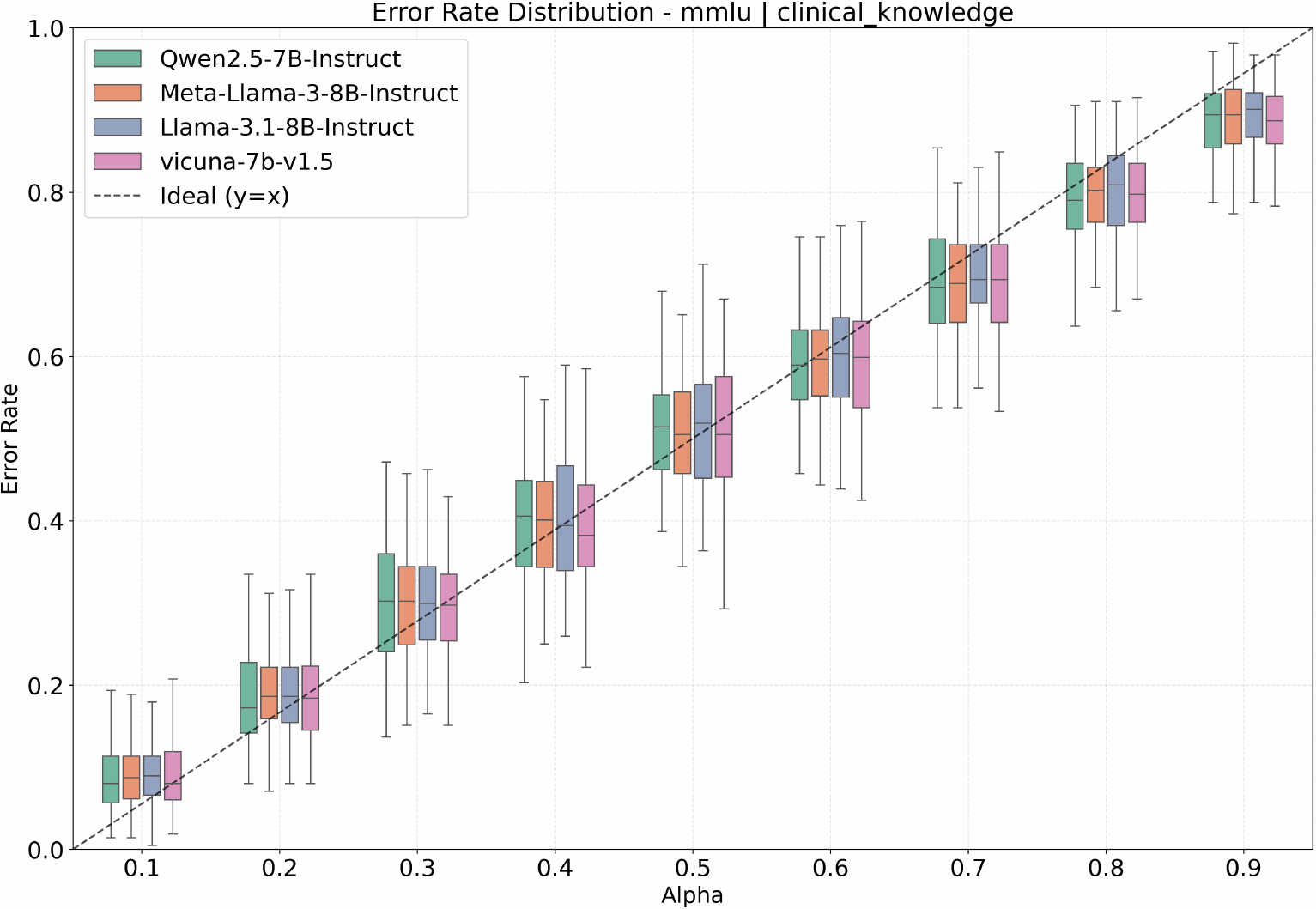}} &
\subfloat[]{\includegraphics[width=5\textwidth]{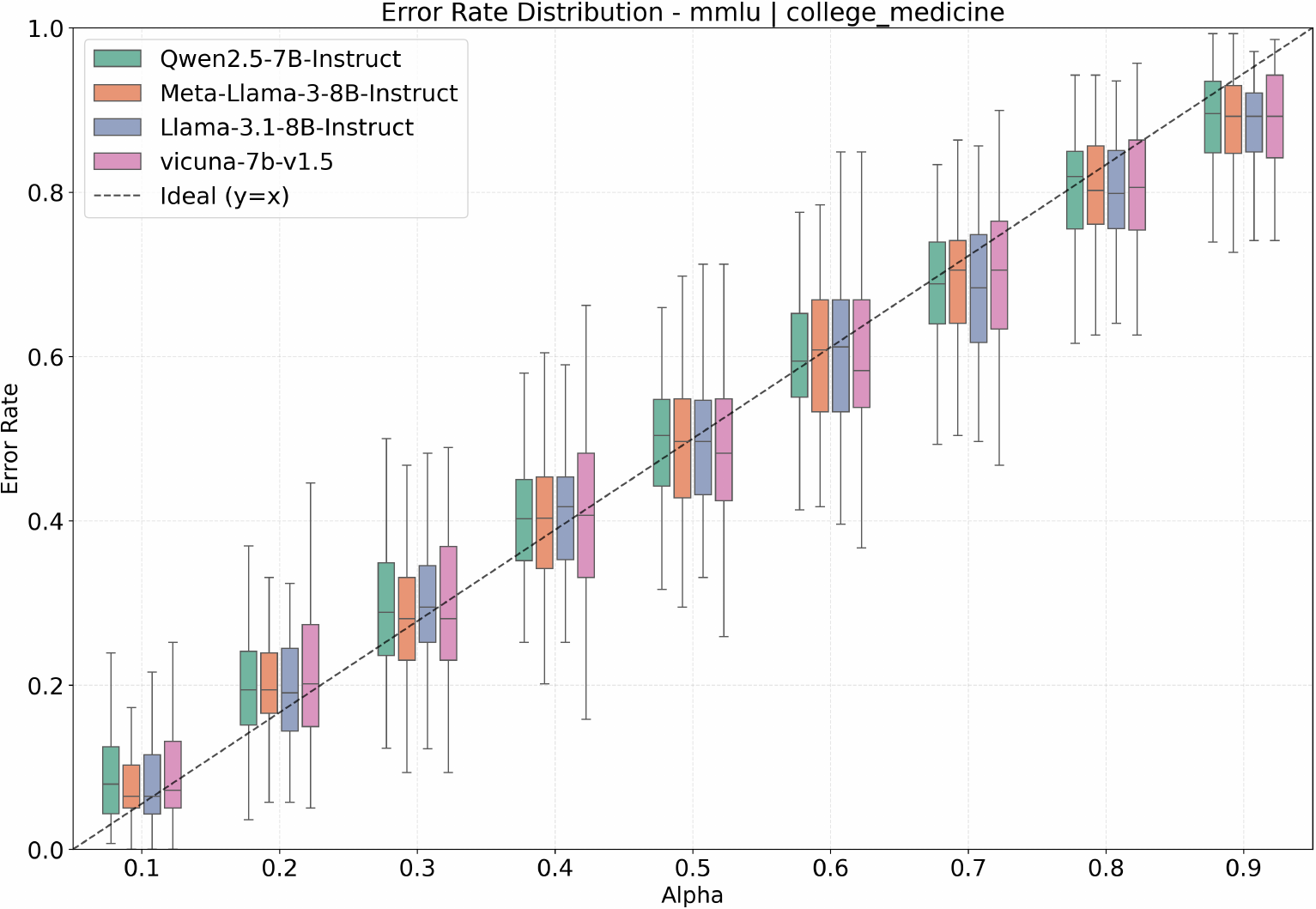}} &
\subfloat[]{\includegraphics[width=5\textwidth]{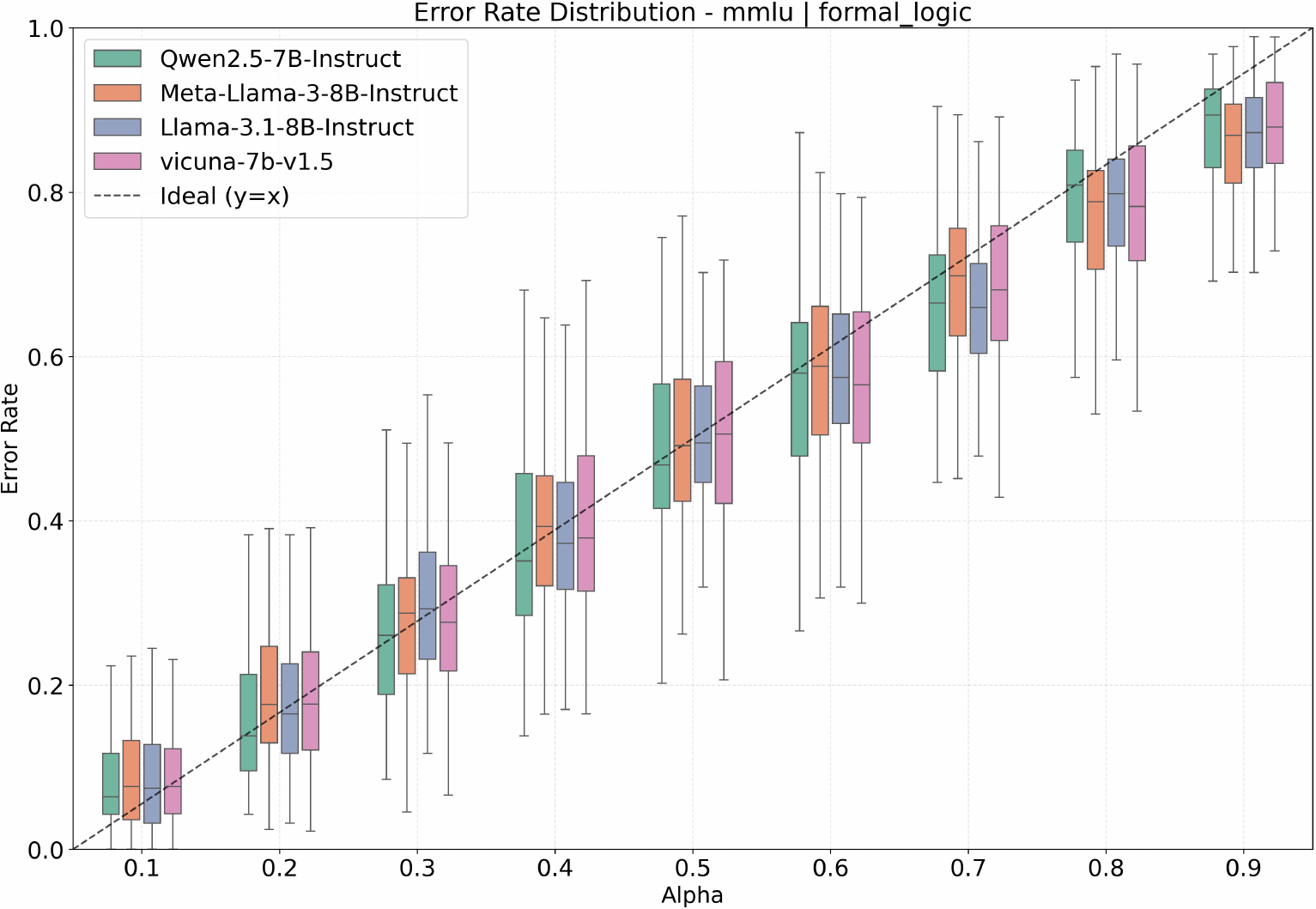}} &
\subfloat[]{\includegraphics[width=5\textwidth]{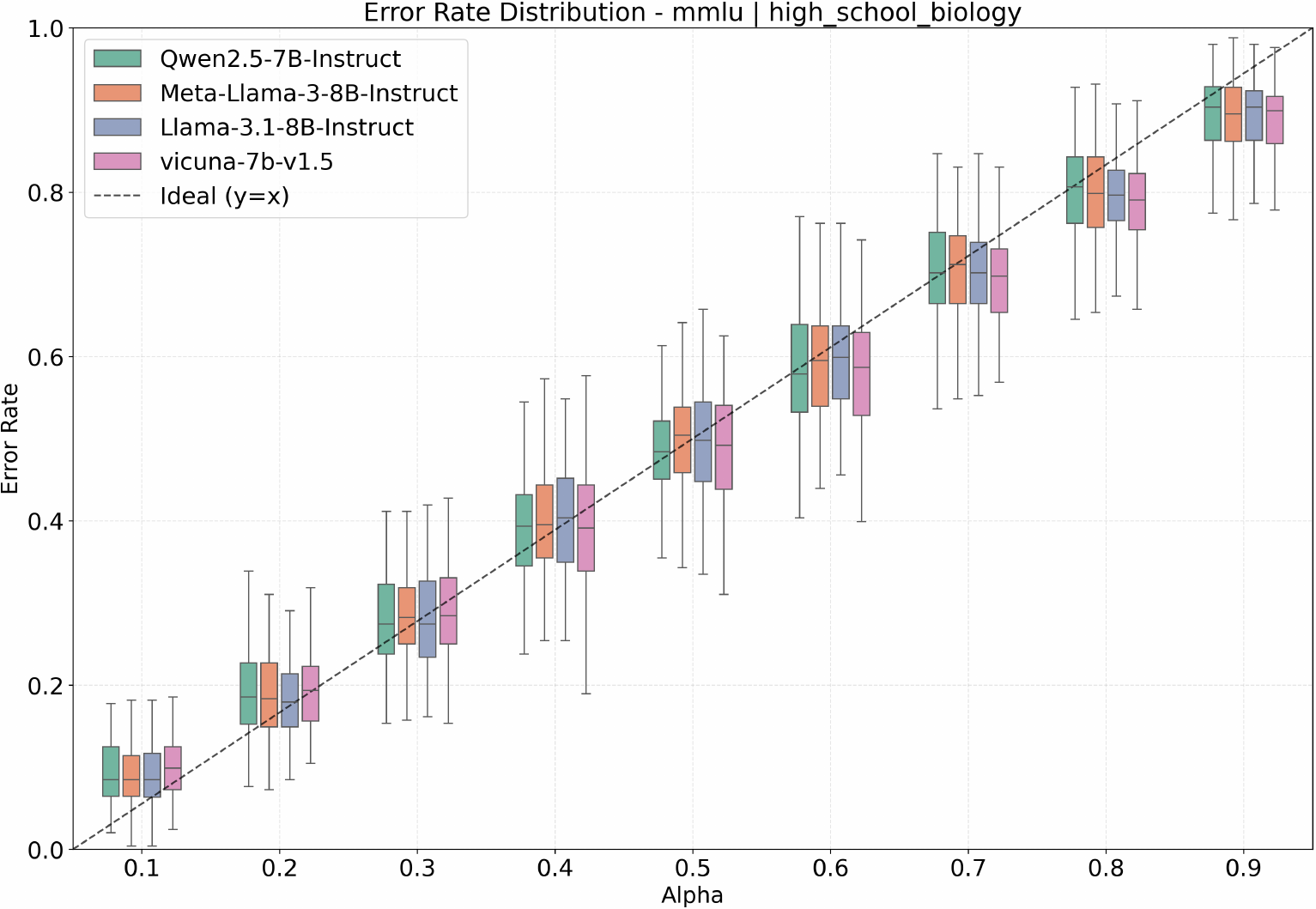}} \\
\subfloat[]{\includegraphics[width=5\textwidth]{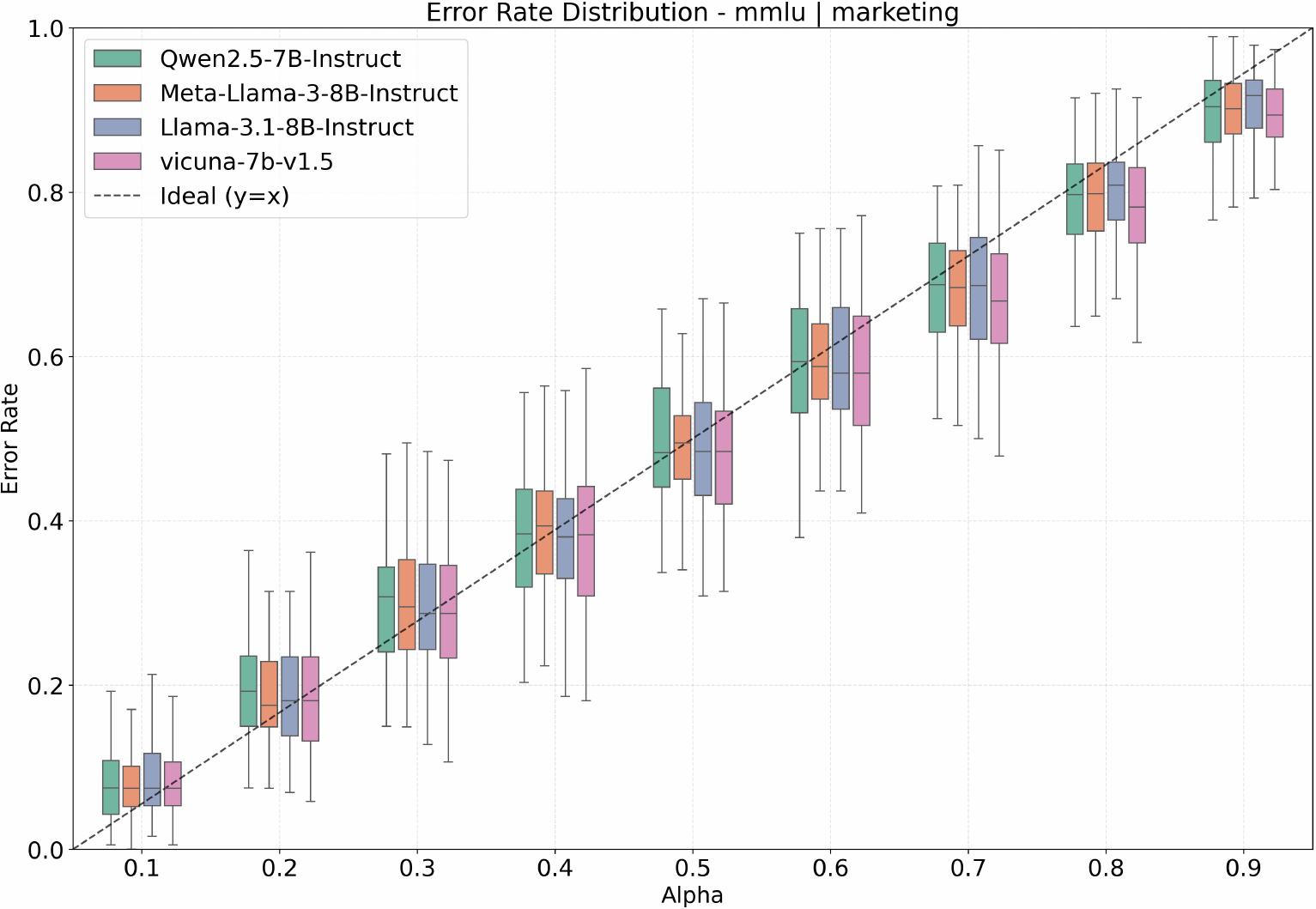}} &
\subfloat[]{\includegraphics[width=5\textwidth]{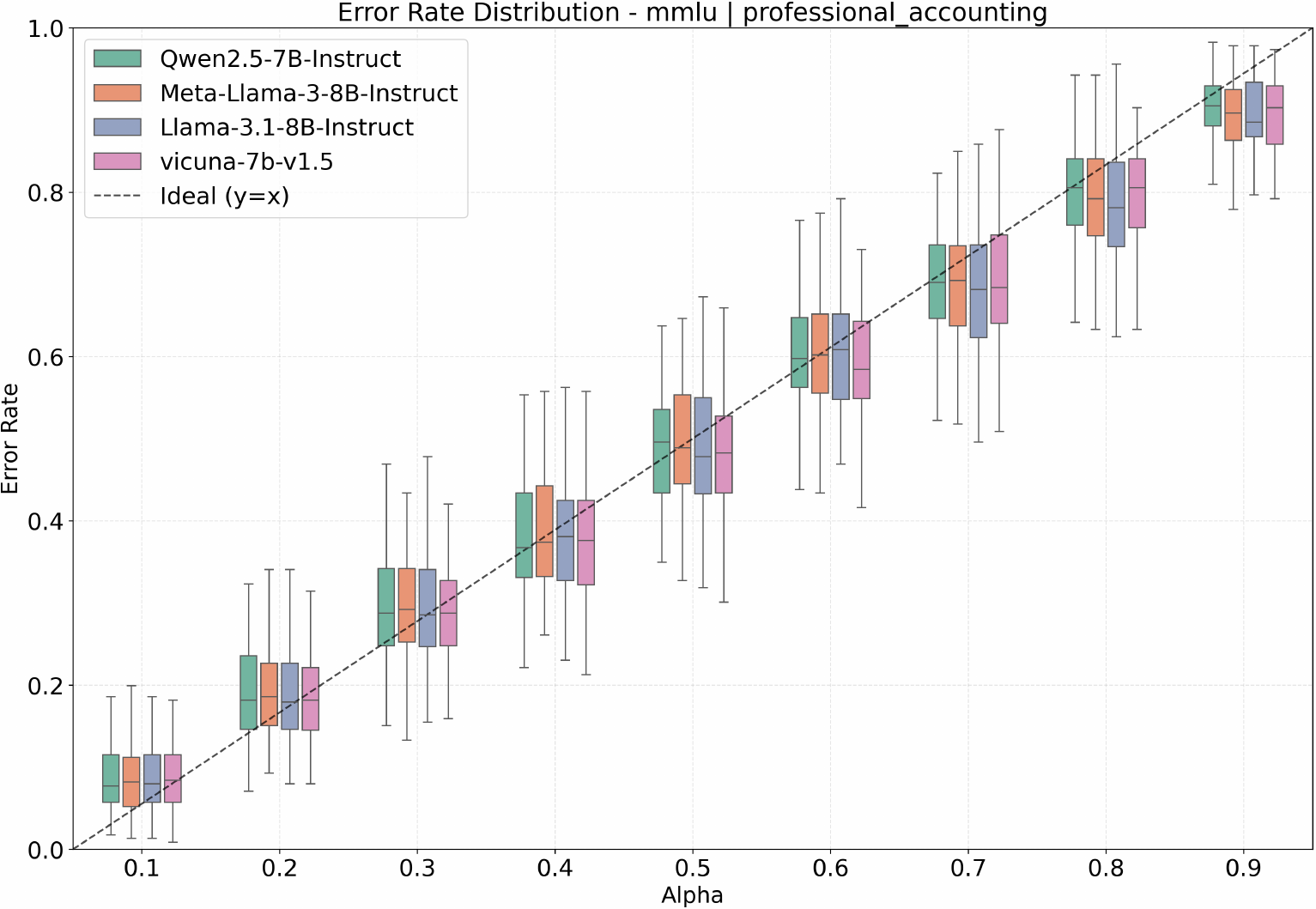}} &
\subfloat[]{\includegraphics[width=5\textwidth]{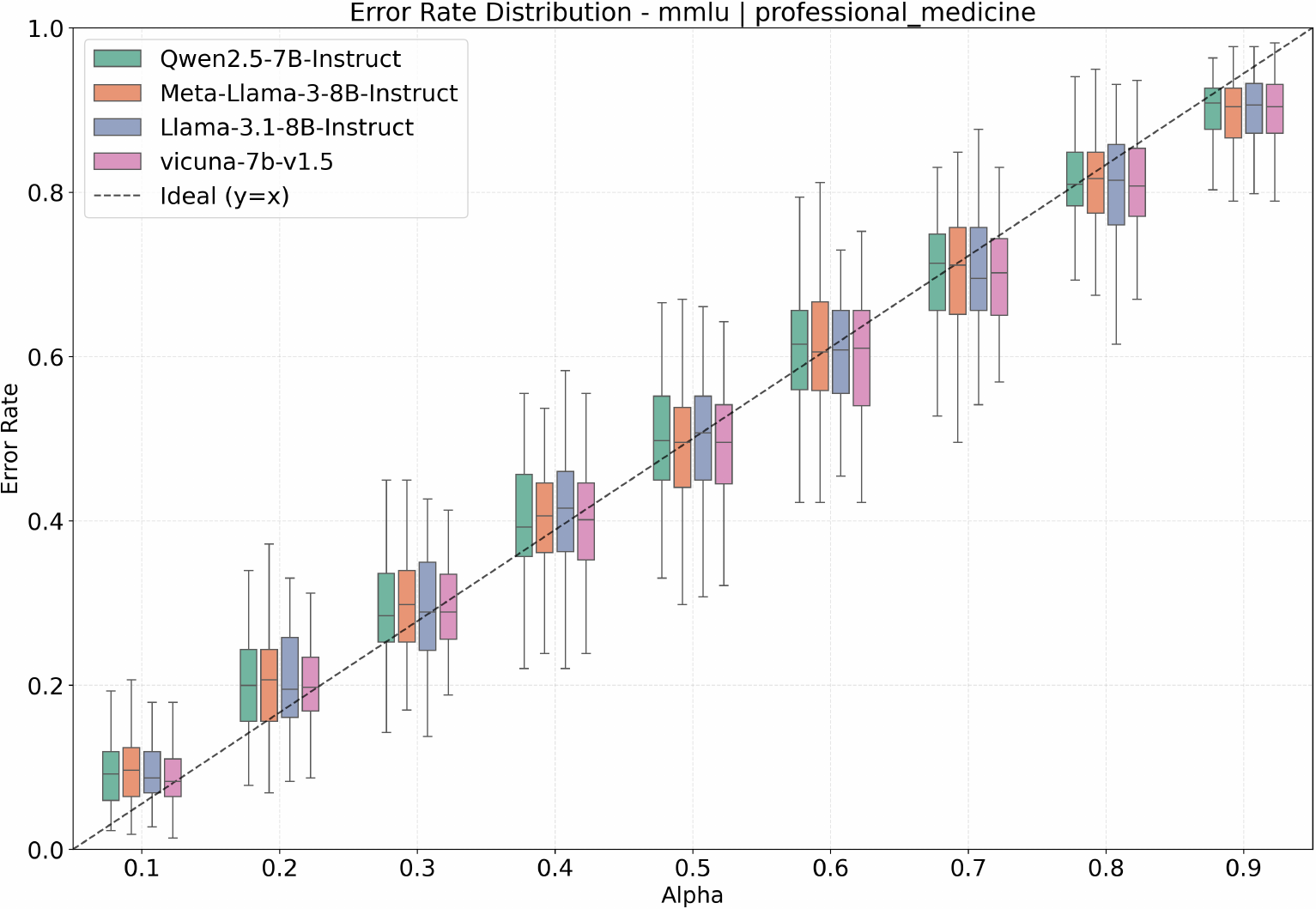}} &
\subfloat[]{\includegraphics[width=5\textwidth]{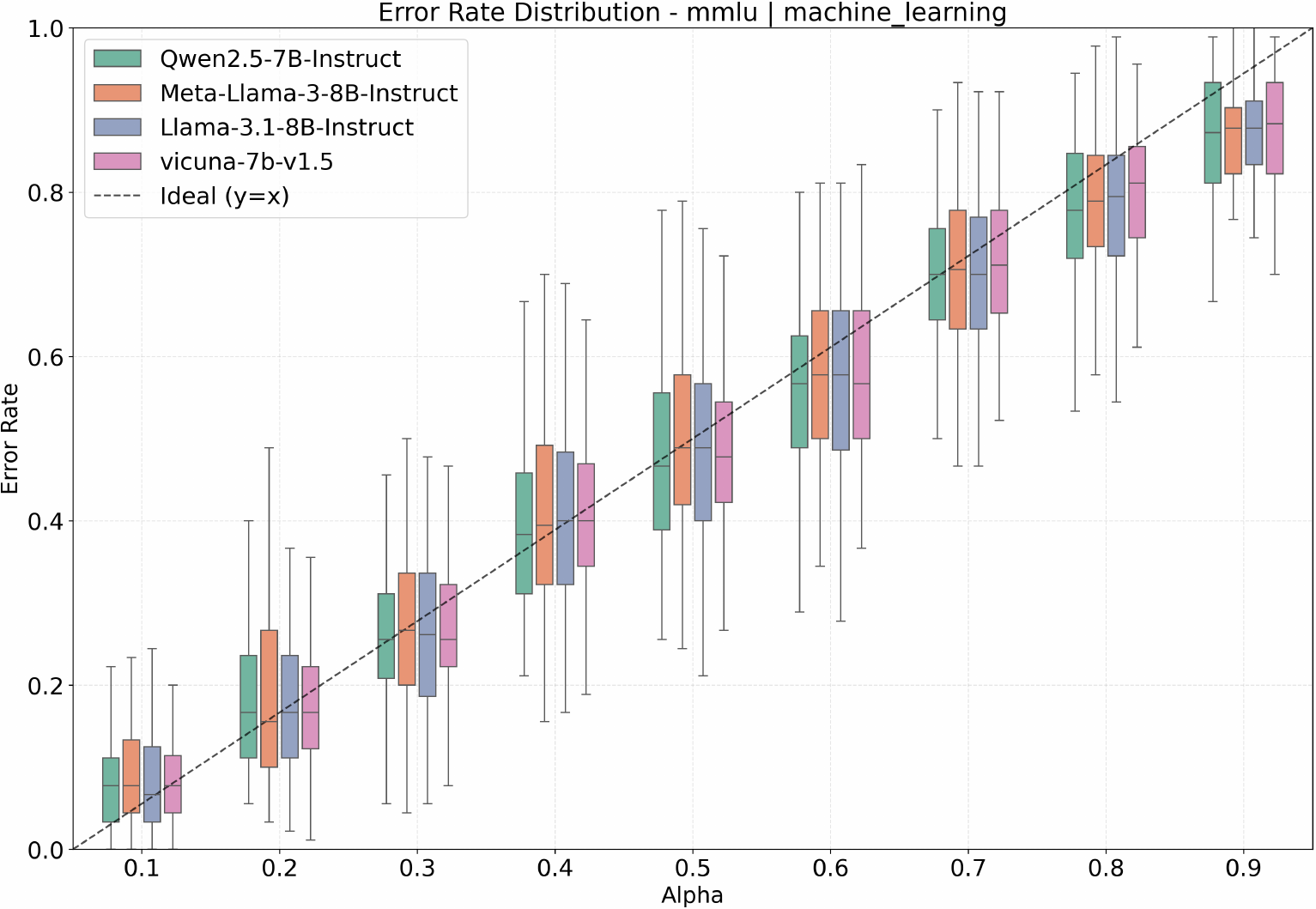}} \\
\end{tabular}
\end{adjustbox}

\caption{Box plots of error rate distribution for 8 subjects with smaller interquartile ranges selected from the MMLU benchmark}
\label{fig3}
\end{figure}

\begin{figure}[!t]
\centering 

\begin{adjustbox}{width=\textwidth}
\begin{tabular}{cccc}
\subfloat[]{\includegraphics[width=5\textwidth]{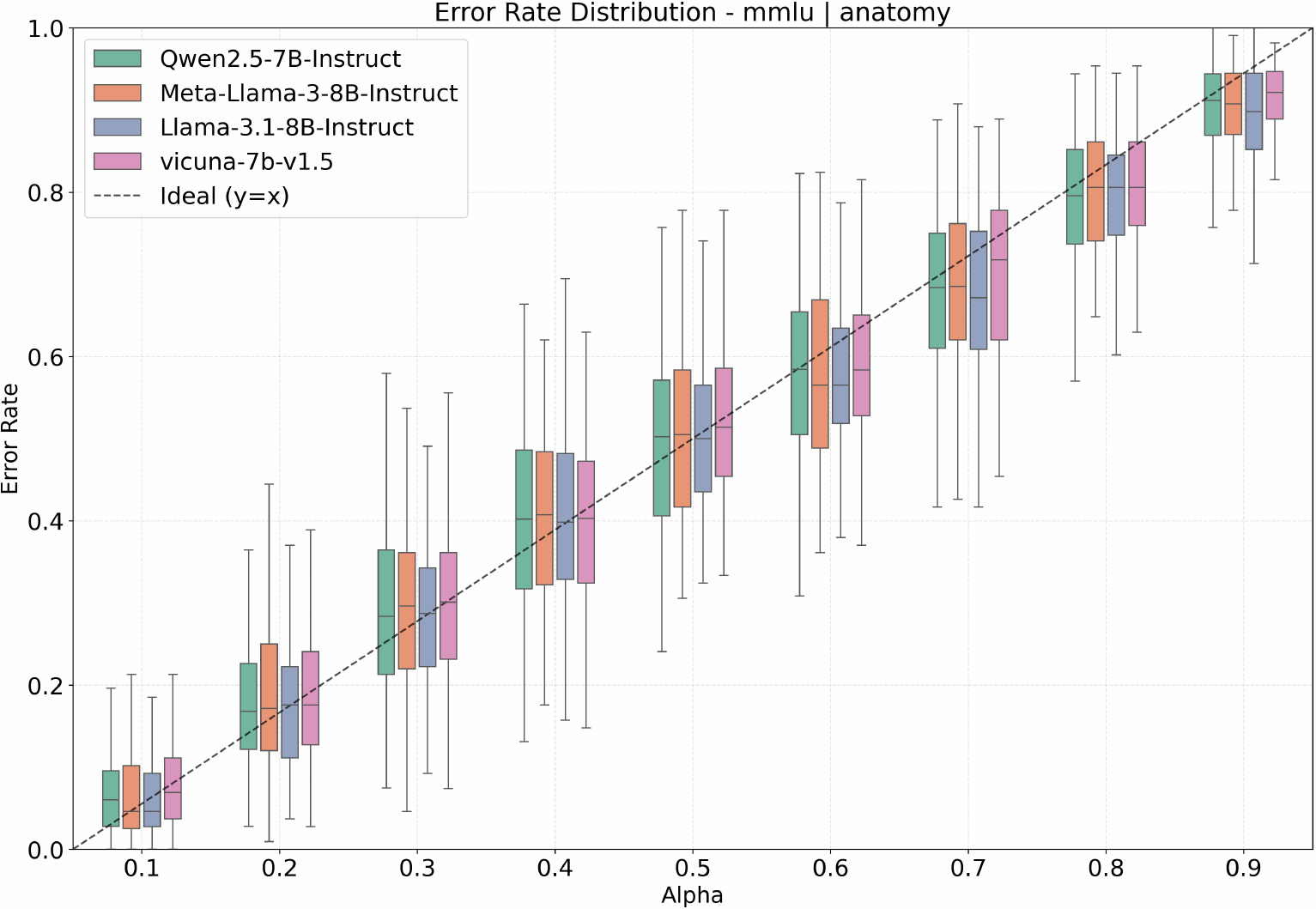}} &
\subfloat[]{\includegraphics[width=5\textwidth]{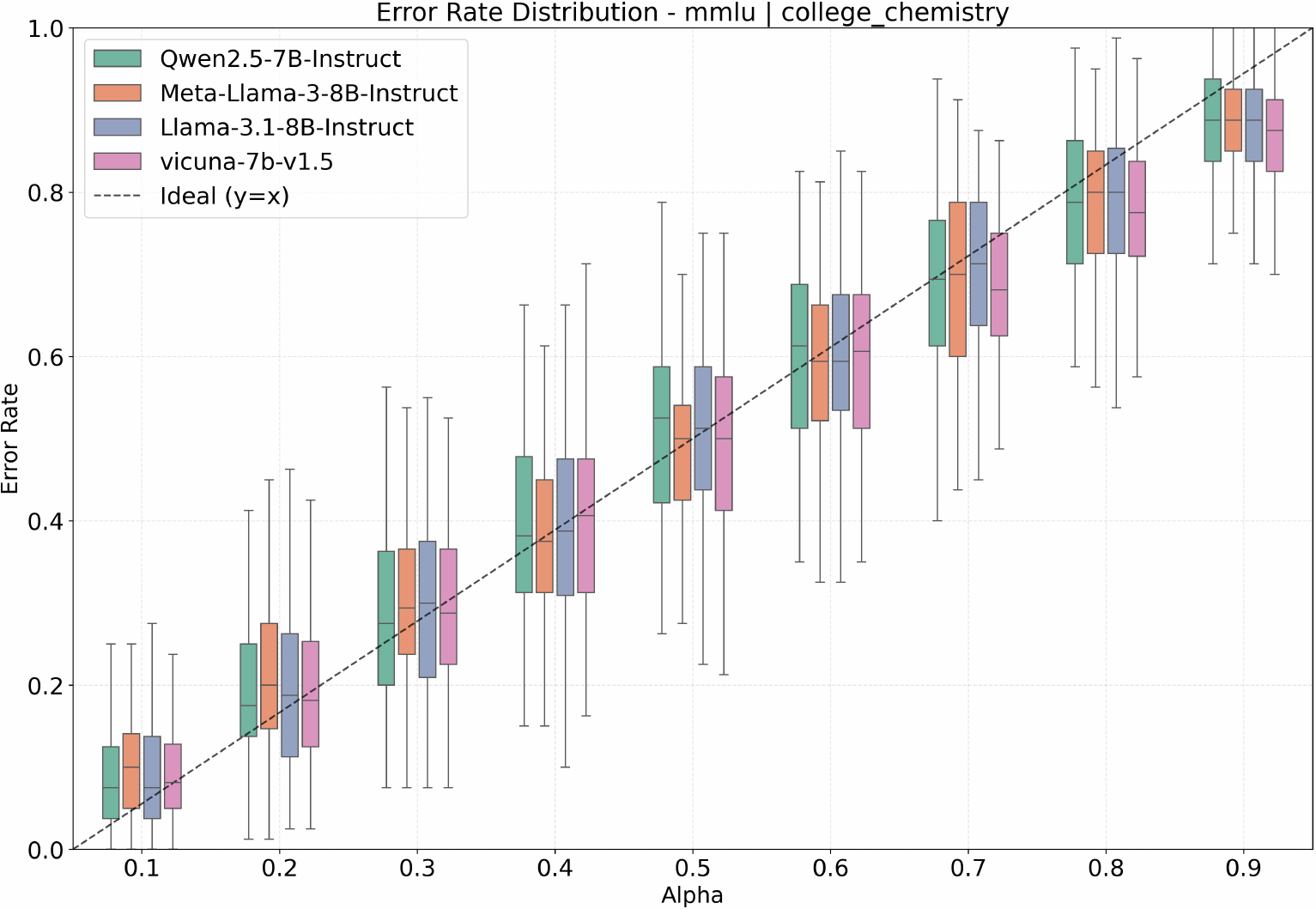}} &
\subfloat[]{\includegraphics[width=5\textwidth]{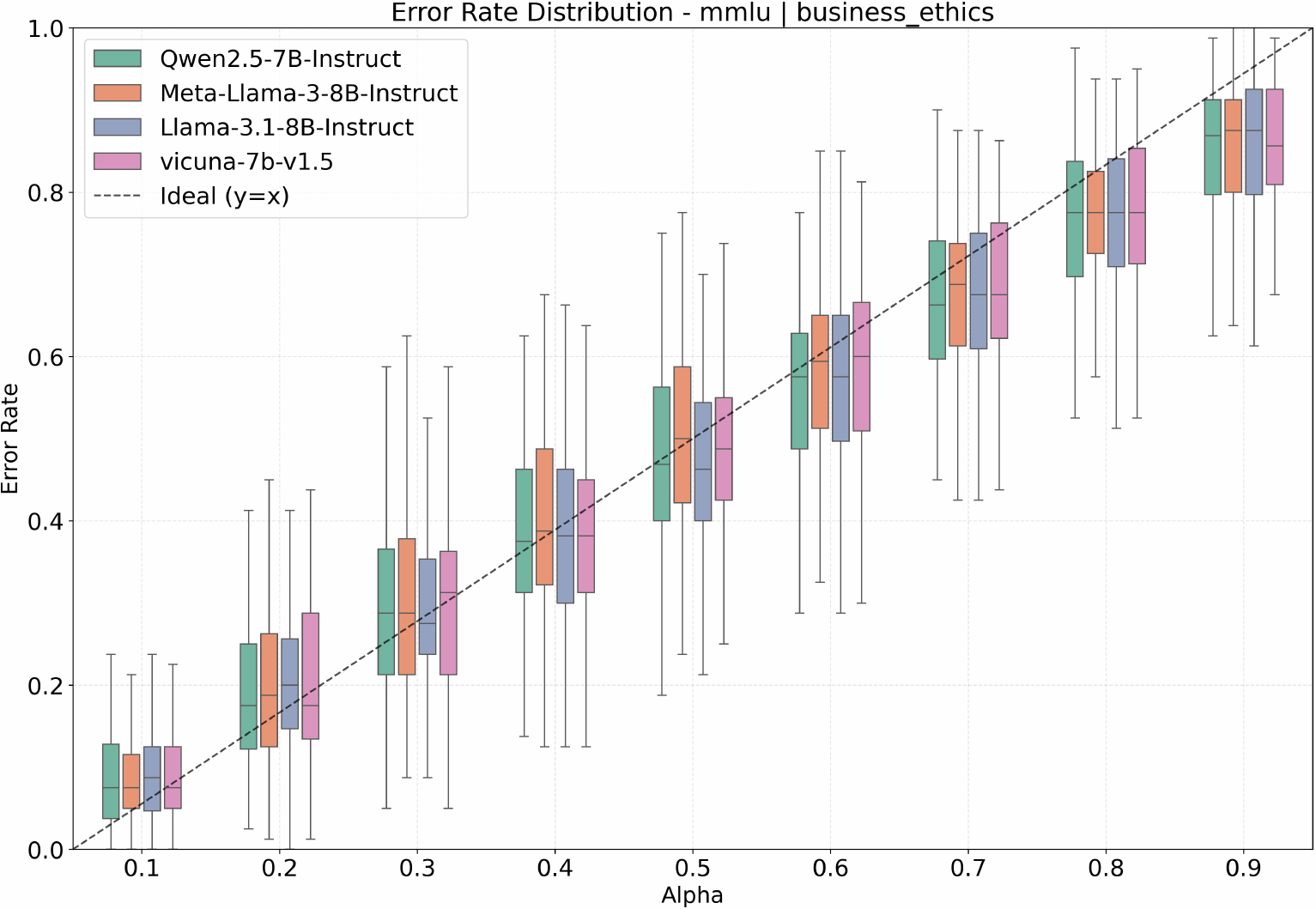}} &
\subfloat[]{\includegraphics[width=5\textwidth]{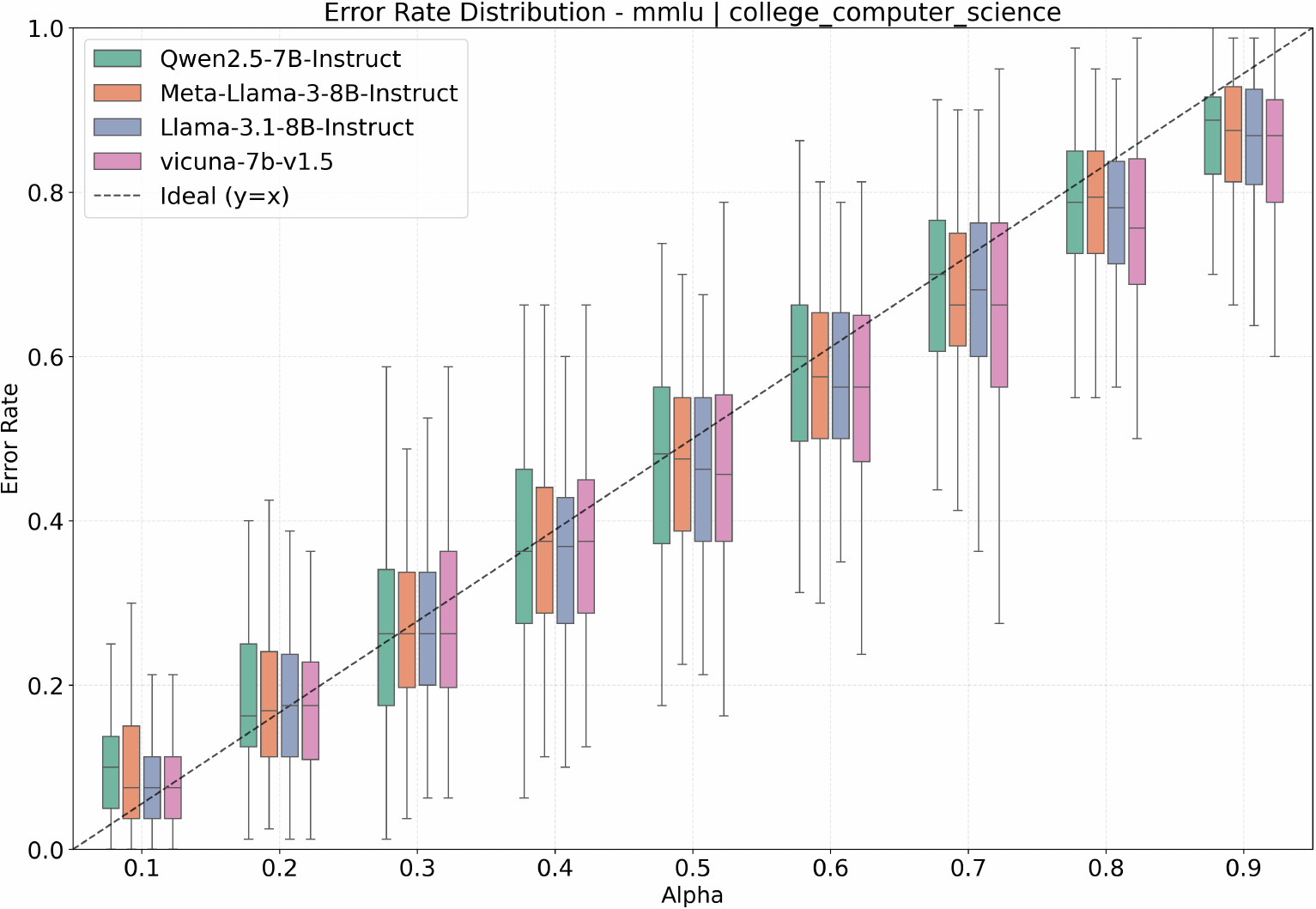}} \\
\subfloat[]{\includegraphics[width=5\textwidth]{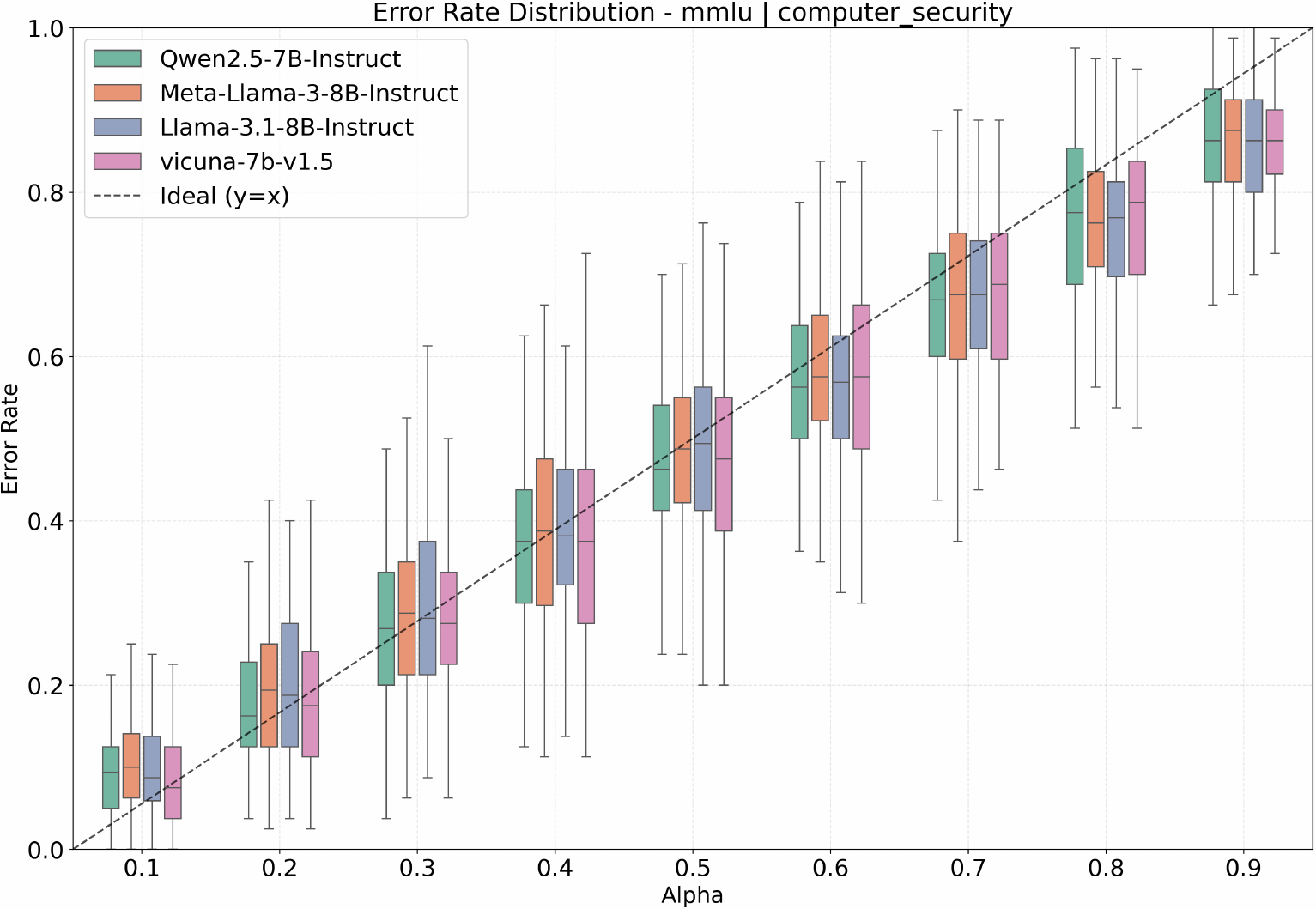}} &
\subfloat[]{\includegraphics[width=5\textwidth]{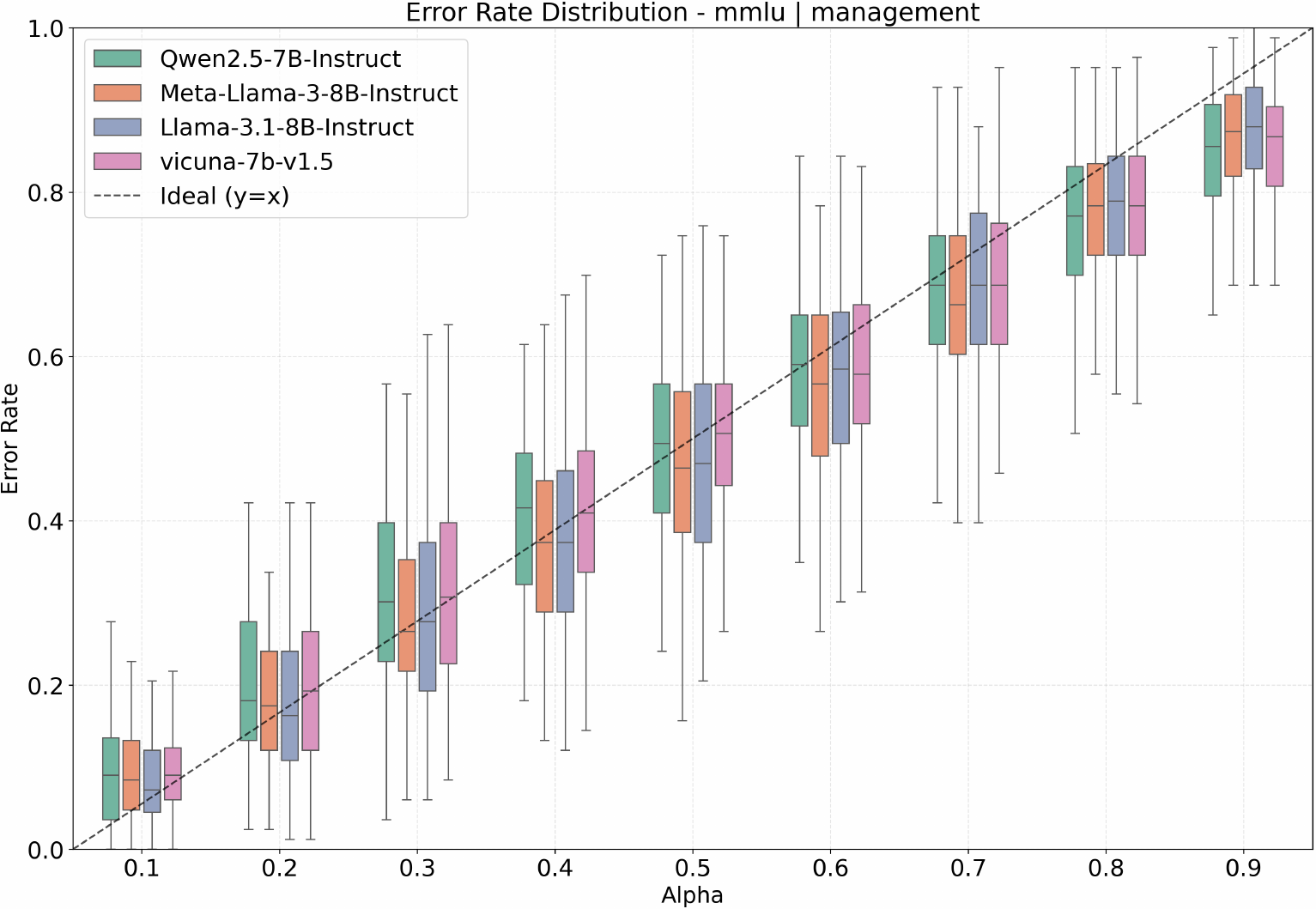}} &
\subfloat[]{\includegraphics[width=5\textwidth]{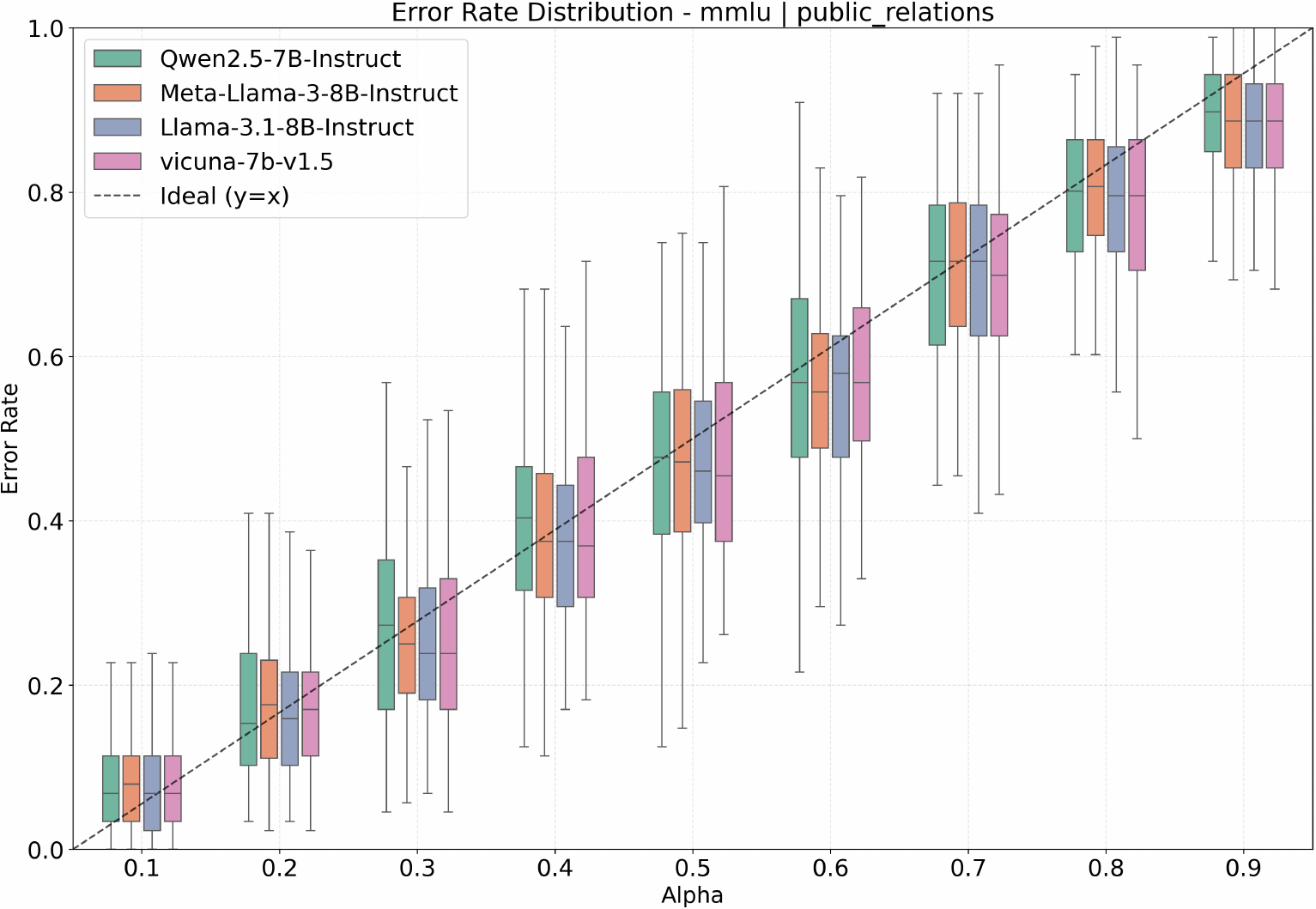}} &
\subfloat[]{\includegraphics[width=5\textwidth]{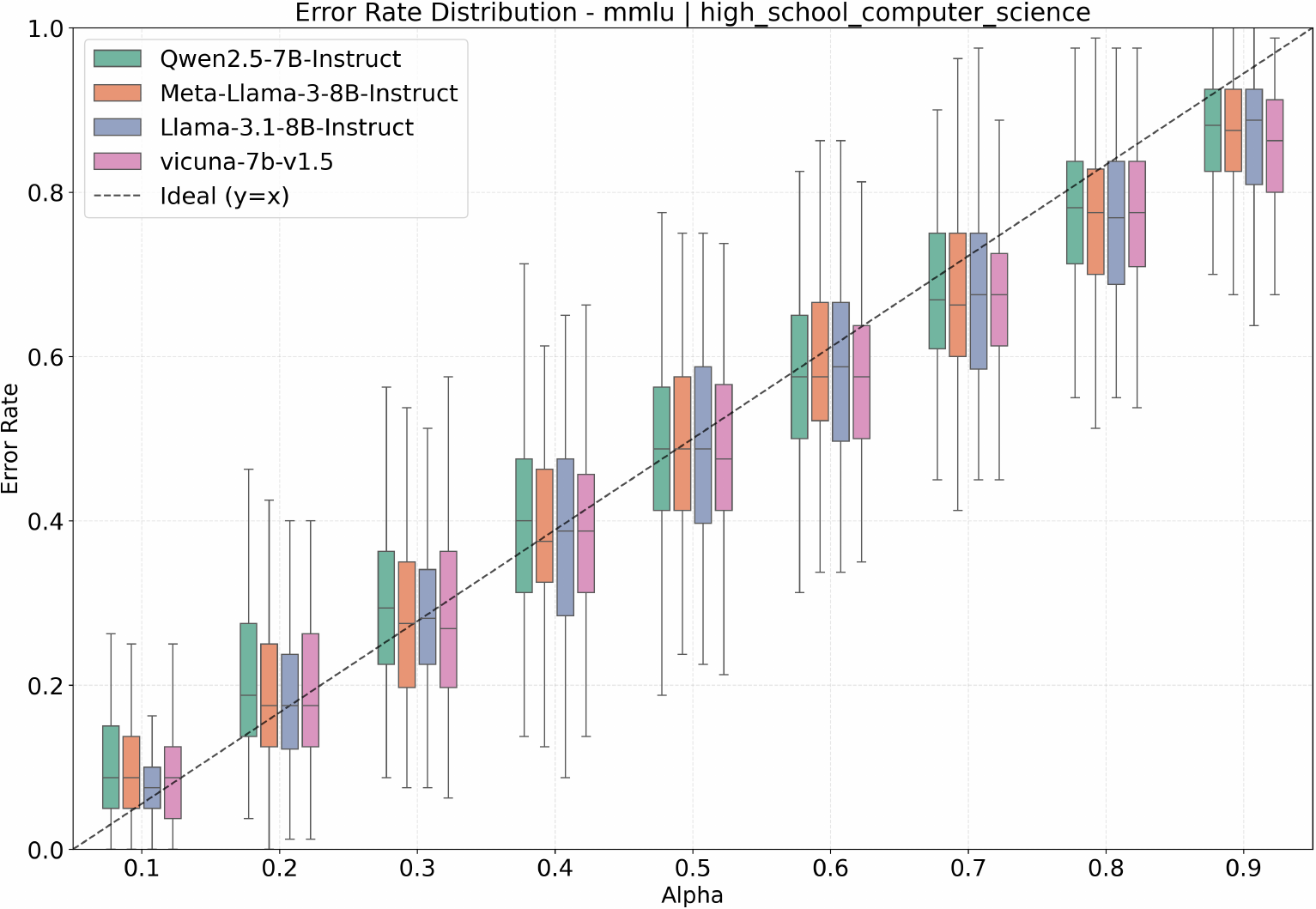}} \\
\end{tabular}
\end{adjustbox}

\caption{Box plots of error rate distribution for 8 subjects with higher interquartile ranges selected from the MMLU benchmark}
\label{fig2}
\end{figure}

Error rate distributions across confidence levels are analyzed for eight minimal-IQR MMLU subtasks in Figure 2. Each subtask visualization plots the confidence parameter against error rate, depicting four LLM variants through box plots that represent interquartile ranges, medians, and data ranges. The theoretical equivalence between confidence parameter and error rate is indicated for reference.

Median error rates decrease monotonically with increasing $\alpha$ across all models, indicating an inverse $\alpha$ error rate association. IQR analysis reveals reduced variability at extreme $\alpha$ values (e.g., 0.1, 0.9) but significantly higher dispersion at intermediate values (e.g., 0.4–0.6), suggesting greater error rate consistency under high-confidence predictions. Most notably, Qwen2.5-3B-Instruct demonstrates the strongest performance: it achieves the lowest median error rate with approximately 75\% of observations below their target alpha values, indicating enhanced accuracy and stability in the evaluated subtasks.target threshold.

Box plots depicting error rate distributions for eight high-dispersion MMLU subjects reveal trends in empirical error rates similar to those observed in minimal-IQR cases. These subjects, however, demonstrate substantially greater variability. Comparative analysis of value ranges at each $\alpha$ level shows more extensive error rate fluctuations across experimental trials, indicating lower model robustness.Nevertheless, Vicuna-7B-v1.5 demonstrates superior performance in these subject-specific tasks, evidenced by both lower median error rates and reduced interquartile ranges.

\begin{figure}[!t]
\centering 

\begin{adjustbox}{width=\textwidth}
\begin{tabular}{ccc}
\subfloat[]{\includegraphics[width=5\textwidth]{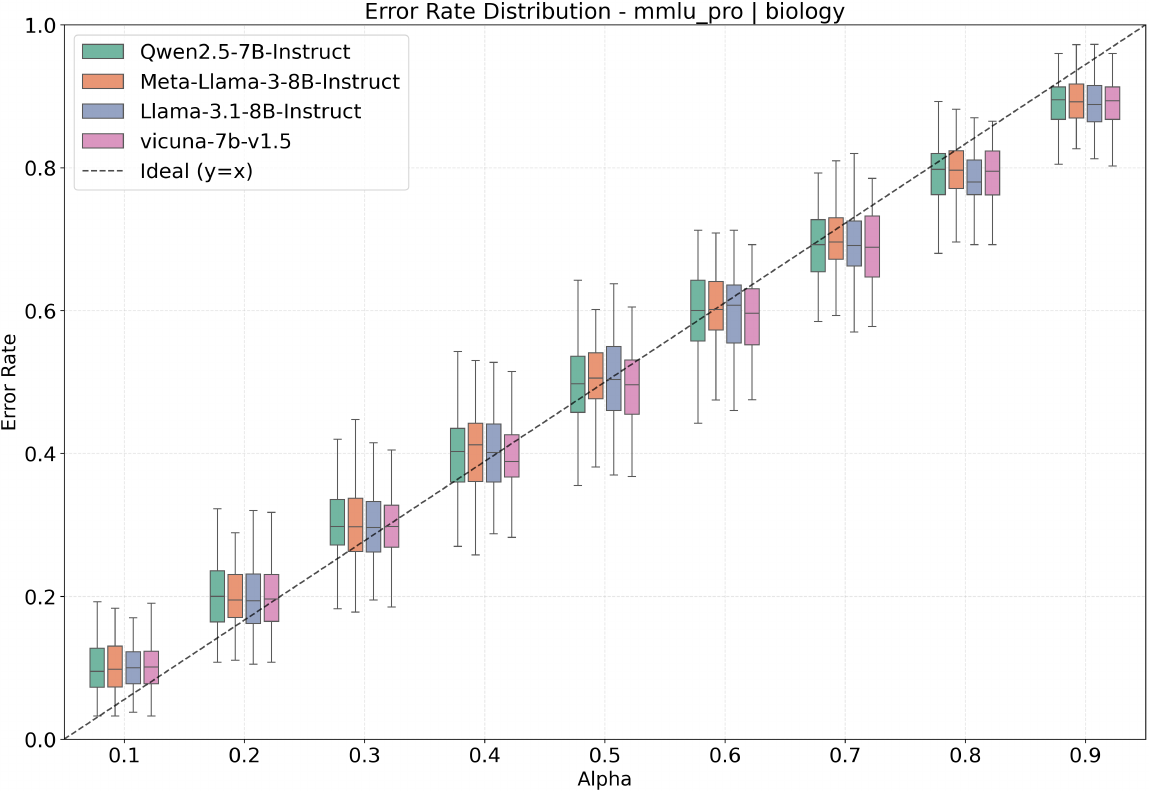}} &
\subfloat[]{\includegraphics[width=5\textwidth]{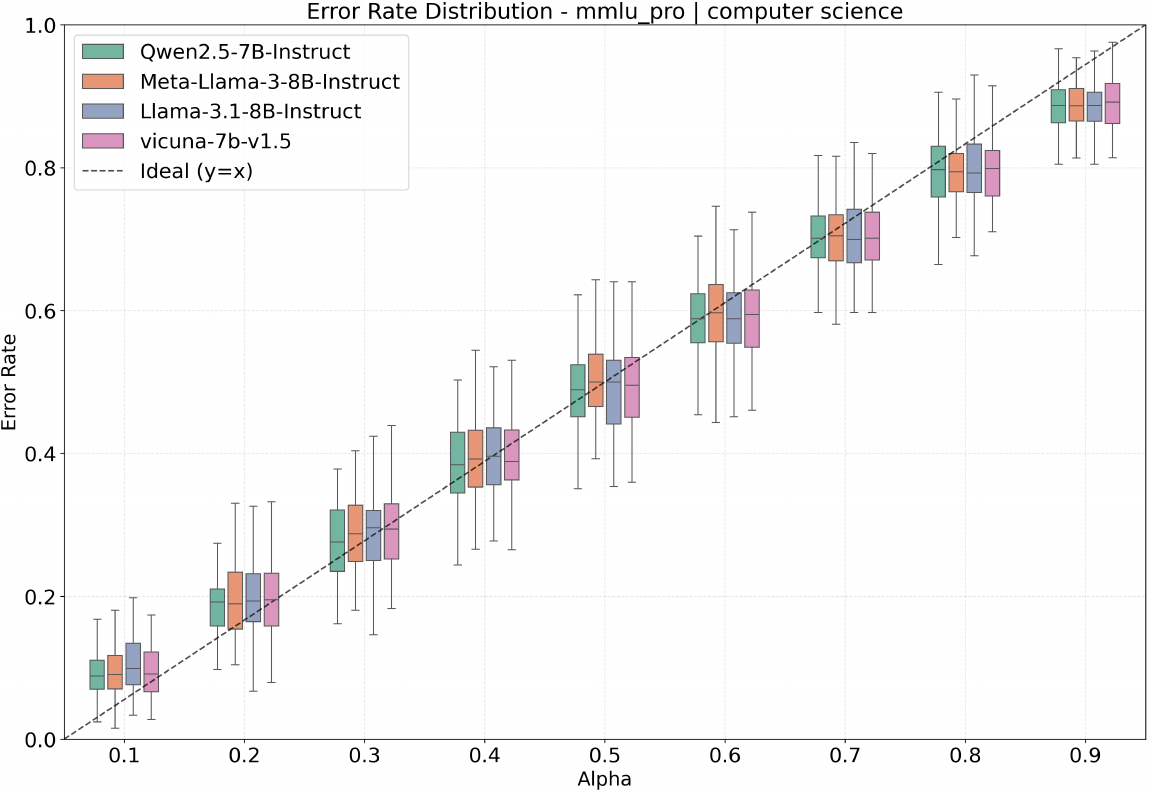}} &
\subfloat[]{\includegraphics[width=5\textwidth]{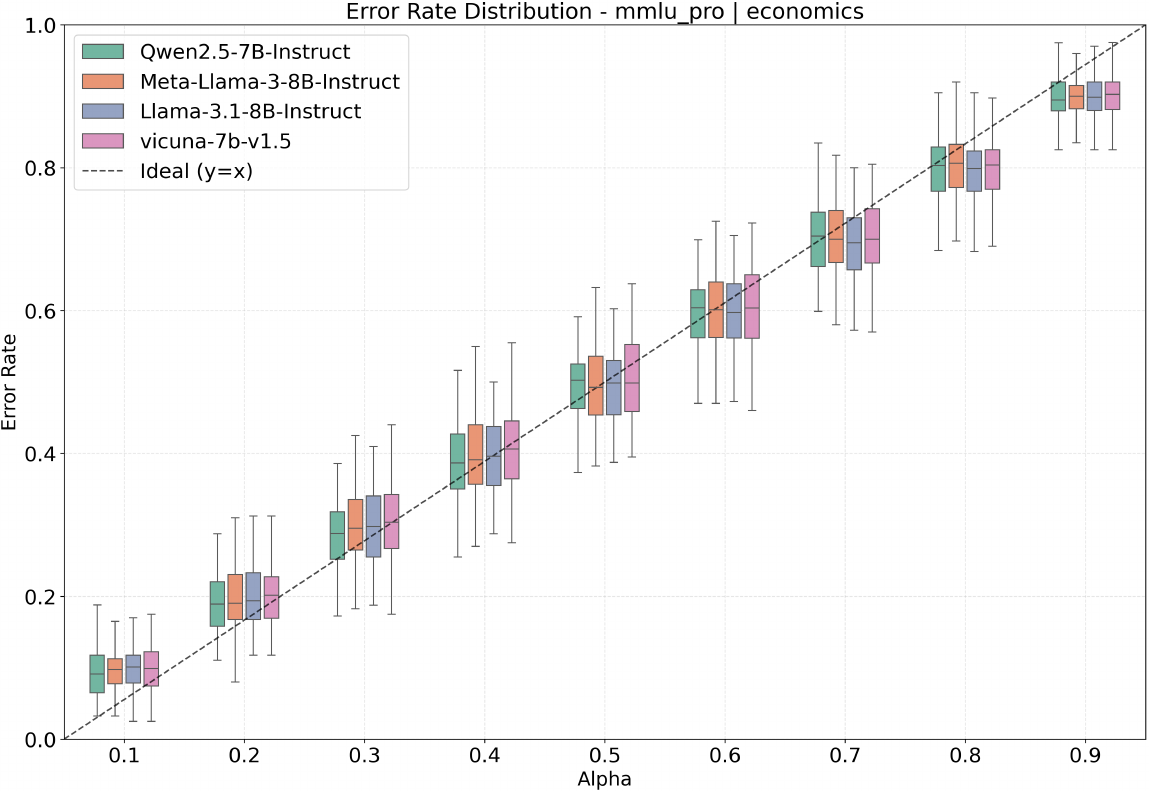}}  \\
\subfloat[]{\includegraphics[width=5\textwidth]{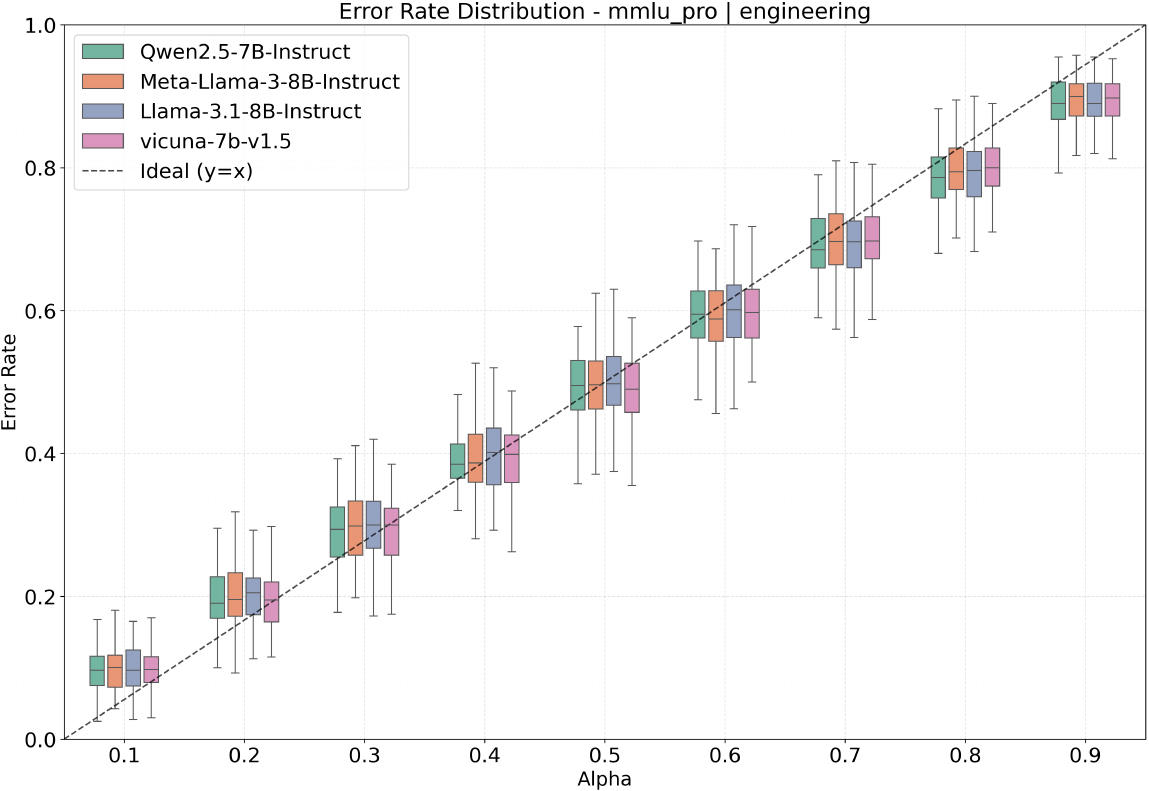}} &
\subfloat[]{\includegraphics[width=5\textwidth]{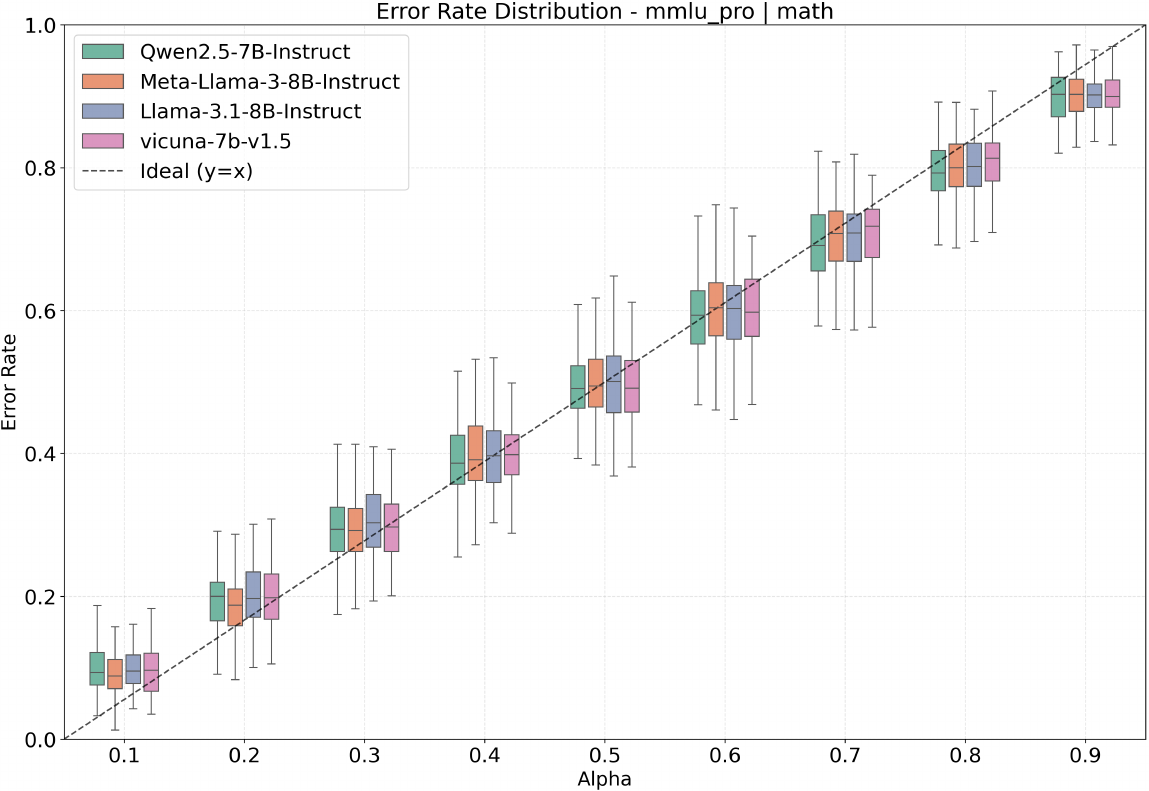}} &
\subfloat[]{\includegraphics[width=5\textwidth]{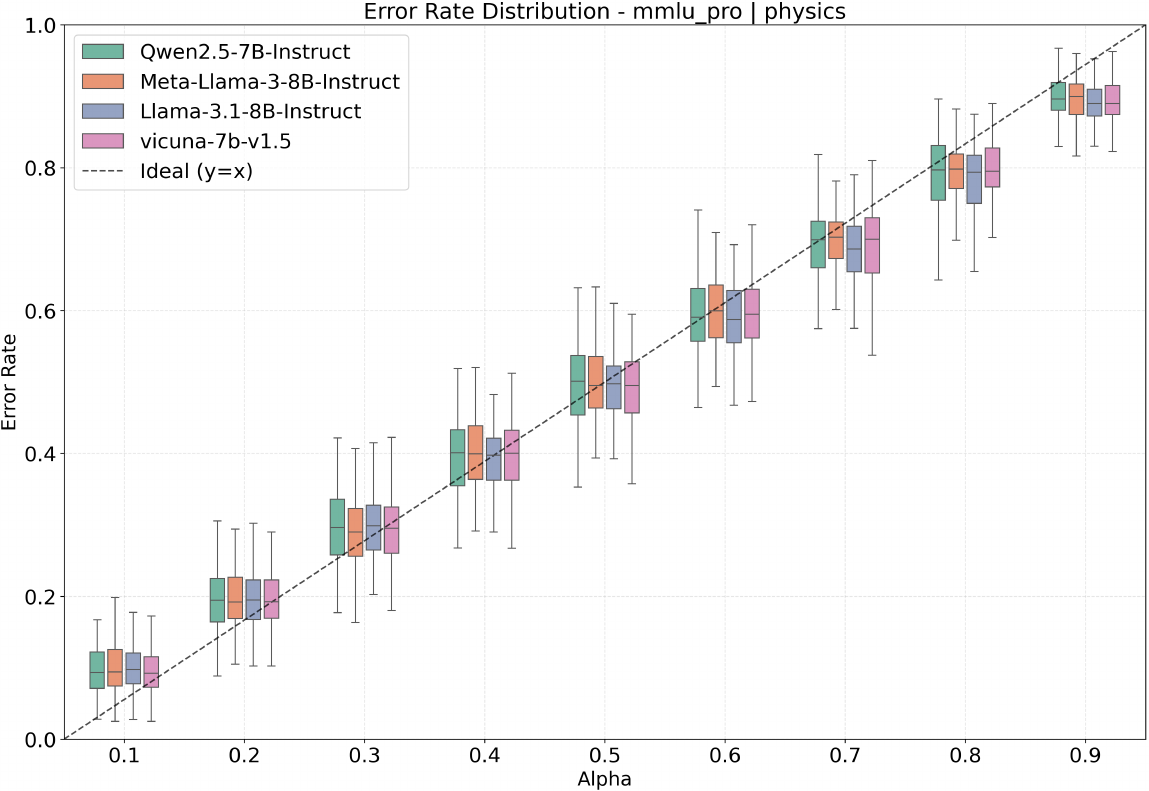}} \\
\end{tabular}
\end{adjustbox}

\caption{Box plots of error rate distribution for 6 subjects  from the MMLU-PRO benchmark}
\label{fig4}
\end{figure}

Given high inter-model result similarity in MMLU-Pro, six representative subjects are presented. Despite increased option complexity and difficulty, models maintain stable empirical error rates. Differences in interquartile ranges between extreme and intermediate $\alpha$ values are less pronounced. Empirical error rates consistently decrease with increasing $\alpha$, while value ranges at identical $\alpha$ levels show significant reduction, indicating enhanced dataset stability. Beyond Qwen2.5-3B-Instruct and Vicuna-7b-v1.5's sustained performance, Llama-3.1-8B-Instruct achieves median error rates comparable to leading models.

\subsection{Prediction Set Size}
This study evaluates the model's impact on prediction set size from two distinct perspectives: the set size for individual subjects and the average set size across datasets. Both metrics were tested on two separate datasets.

\begin{table}[!t]
\centering
\caption{Result of the prediction set size at various risk levels}
\begin{tabular}{lllllllllll} 
\hline
Dataset                   & LLMs/$\alpha$ & 0.1  & 0.2  & 0.3  & 0.4  & 0.5  & 0.6  & 0.7  & 0.8  & 0.9   \\ 
\hline
\multirow{4}{*}{MMLU}     & Llama-3.1-8B-Instruct          & 1.98 & 1.48 & 1.16 & 0.89 & 0.66 & 0.49 & 0.34 & 0.22 & 0.11  \\
                          & Meta-Llama-3-8B-Instruct       & 2.07 & 1.54 & 1.19 & 0.92 & 0.69 & 0.50 & 0.35 & 0.22 & 0.11  \\
                          & Qwen2.5-7B-Instruct            & 1.98 & 1.41 & 1.07 & 0.83 & 0.64 & 0.47 & 0.33 & 0.22 & 0.11  \\
                          & vicuna-7b-v1.5                 & 2.94 & 2.32 & 1.89 & 1.50 & 1.12 & 0.82 & 0.55 & 0.33 & 0.14  \\ 
\hline
\multirow{4}{*}{MMLU-PRO} & Llama-3.1-8B-Instruct          & 6.00 & 4.51 & 3.33 & 2.44 & 1.73 & 1.20 & 0.78 & 0.44 & 0.21  \\
                          & Meta-Llama-3-8B-Instruct       & 6.78 & 5.12 & 3.82 & 2.78 & 1.95 & 1.28 & 0.79 & 0.44 & 0.20  \\
                          & Qwen2.5-7B-Instruct            & 6.58 & 5.01 & 3.61 & 2.40 & 1.60 & 1.04 & 0.64 & 0.37 & 0.18  \\
                          & vicuna-7b-v1.5                 & 7.86 & 6.70 & 5.53 & 4.43 & 3.39 & 2.44 & 1.62 & 0.92 & 0.41  \\
\hline
\end{tabular}
\end{table}

Tabular data confirm that increasing confidence threshold $\alpha$ monotonically reduces the Average Prediction Set Size (APSS) across all models, consistent with selective prediction theory. Llama-series models demonstrate superior compression efficiency, particularly in high-confidence regimes ($\alpha \geq 0.7$), achieving 33\% smaller prediction sets than Vicuna counterparts at $\alpha=0.8$. In contrast, Qwen2.5 models exhibit gradual reduction patterns, maintaining larger prediction sets at intermediate confidence levels ($\alpha=0.5$). Vicuna models consistently yield the highest APSS values, indicating substantially elevated predictive uncertainty.



\begin{figure}[!t]
\centering 

\begin{adjustbox}{width=\textwidth}
\begin{tabular}{ccc}
\subfloat[]{\includegraphics[width=5\textwidth]{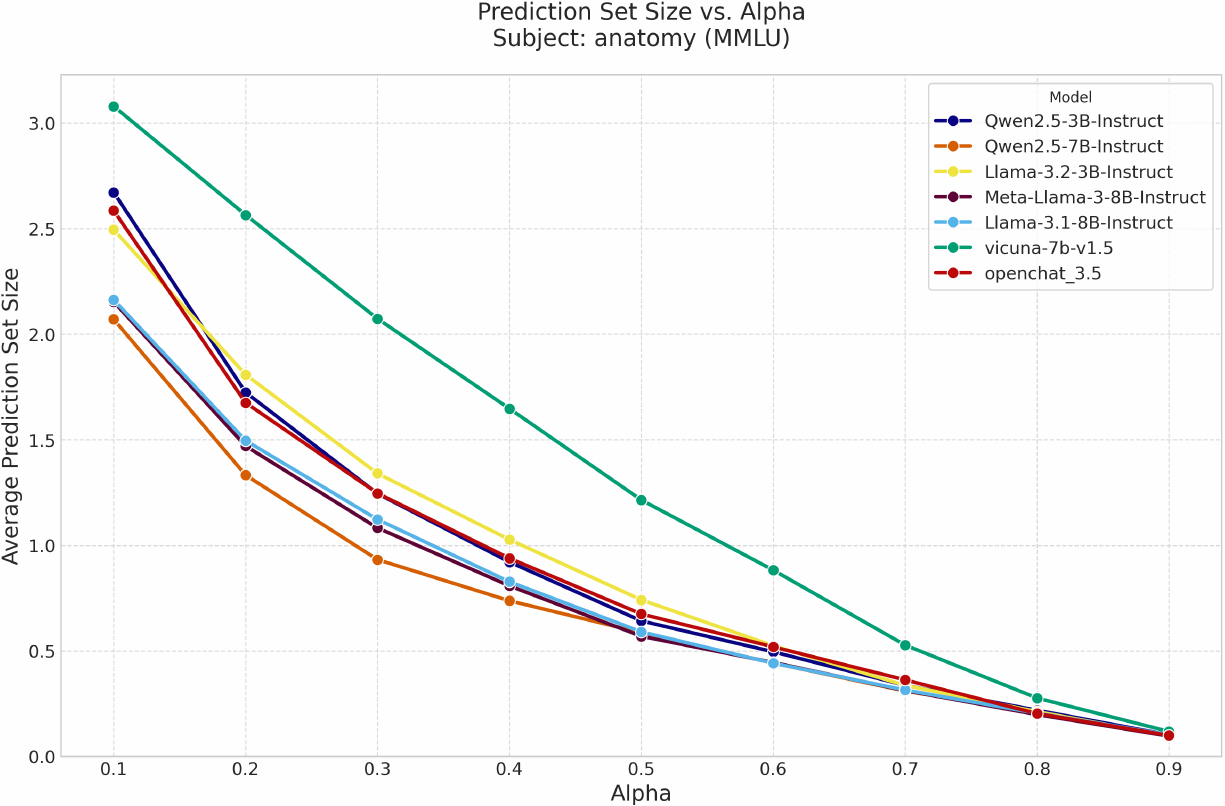}} &
\subfloat[]{\includegraphics[width=5\textwidth]{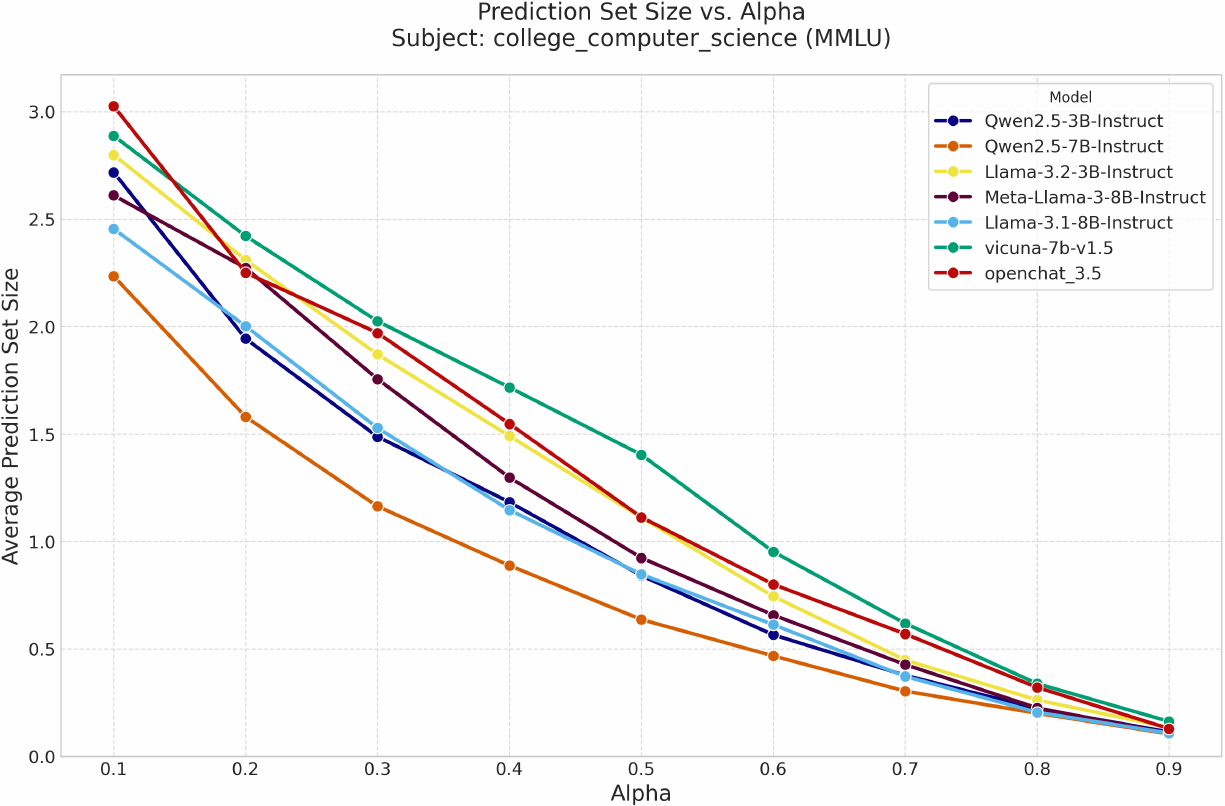}} \\
\subfloat[]{\includegraphics[width=5\textwidth]{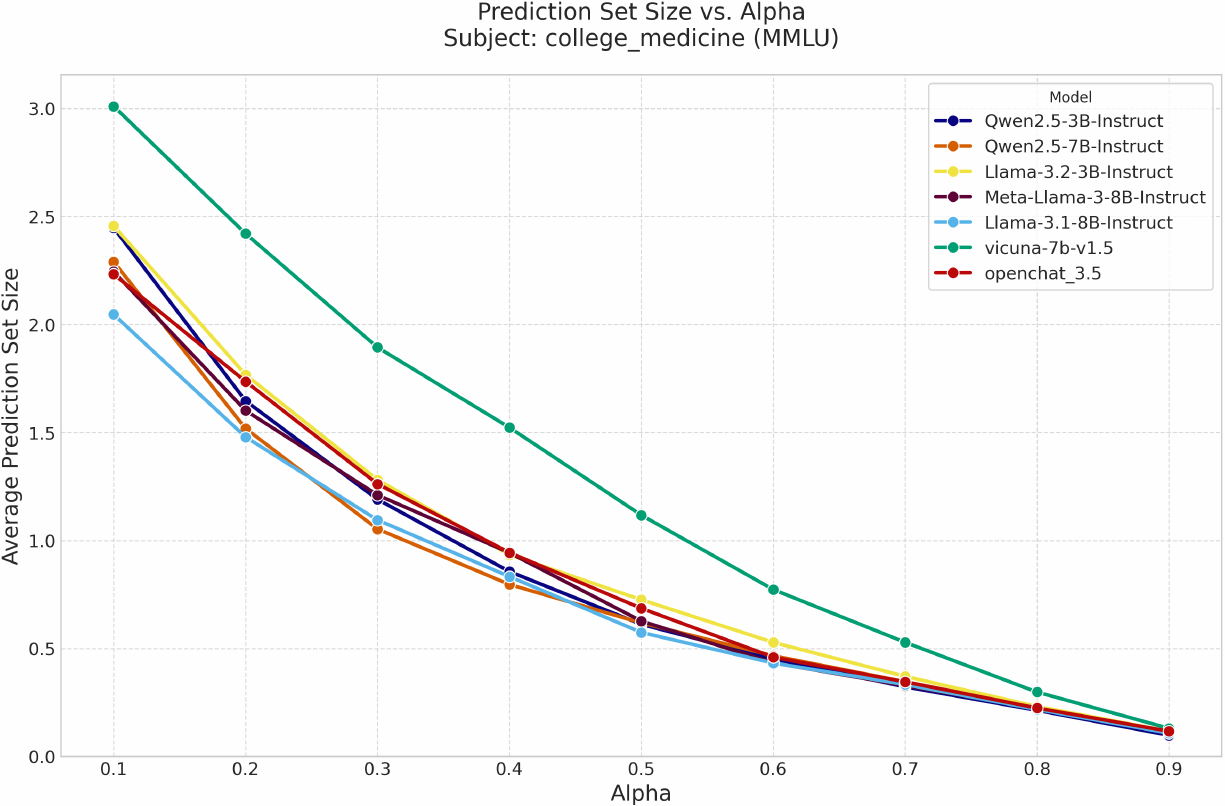}} &
\subfloat[]{\includegraphics[width=5\textwidth]{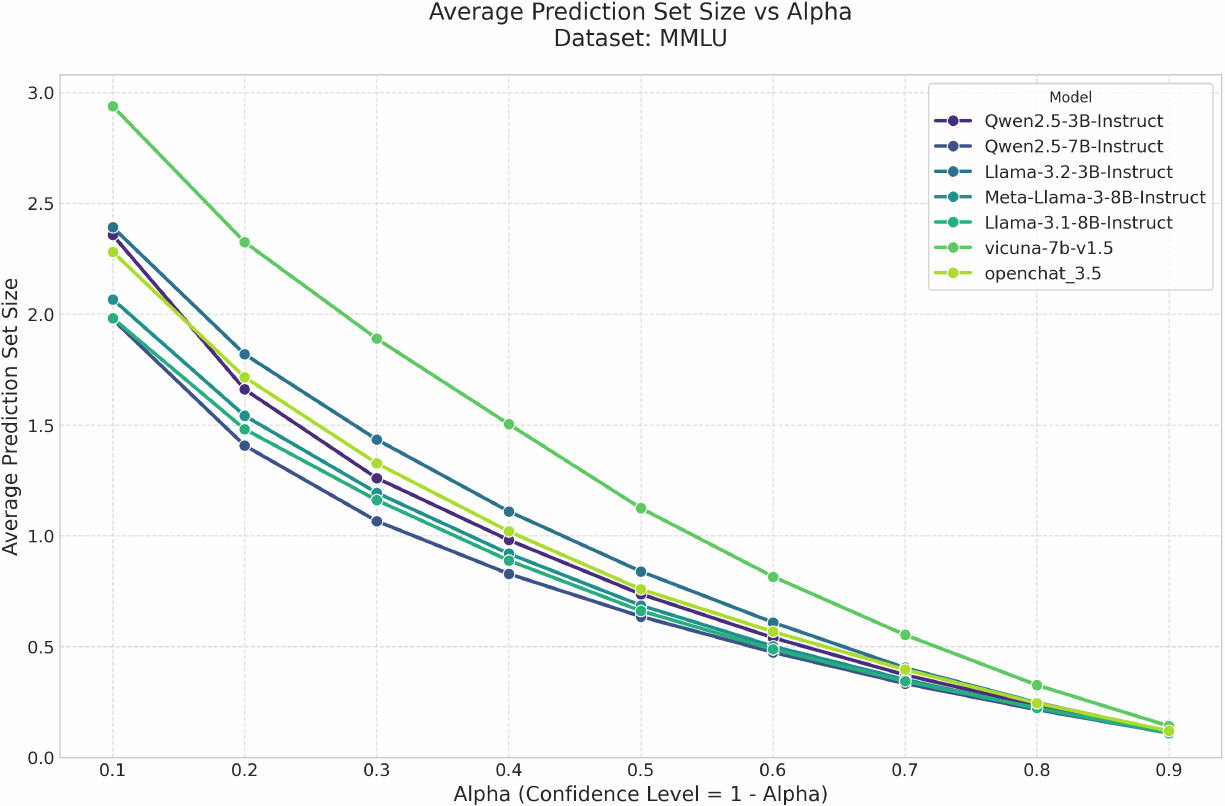}} \\
\end{tabular}
\end{adjustbox}

\caption{Comparative Curves of Individual Subject and Average Prediction Set Sizes in MMLU}

\end{figure}

Dataset complexity systematically modulates uncertainty management. In more challenging benchmarks, all models produce prediction sets exceeding baseline sizes by factors >3. This discrepancy amplifies at higher confidence levels, with maximum inter-model differences reaching 2.3$\times$ at $\alpha=0.6$. Notably, Qwen2.5 demonstrates comparative advantage under stringent conditions ($\alpha=0.9$), suggesting enhanced coverage stability in high-uncertainty environments. This task-dependent behavior necessitates context-specific calibration.

The APSS metric effectively quantifies uncertainty management efficiency. Llama's exceptional compression reflects precise confidence calibration, ideal for efficiency-critical applications. Progressive decay of Qwen2.5 provides more stable risk coverage for reliability-sensitive contexts. These findings establish APSS as a diagnostic tool that reveals calibration limitations and informs model optimization through balanced compression-coverage strategies for risk-stratified decision scenarios.

The variation in prediction set size across the entire MMLU dataset and individual subjects (Fig. 5) reveals three principal patterns. Prediction set size decreases monotonically with increasing significance level $\alpha$, demonstrating $\alpha$'s regulatory role in the confidence-reliability trade-off: higher $\alpha$ values yield higher-confidence predictions with fewer candidate classes, whereas lower $\alpha$ retains more classes to ensure reliability. At $\alpha=0.4$, most models converge to prediction sets of cardinality 1, establishing this threshold as a critical balance point between efficiency and accuracy for practical $\alpha$ configuration. Significant behavioral differences among models are observed: Vicuna-7B-v1.5 maintains larger initial set sizes with near-linear reduction at low $\alpha$ ($\alpha < 0.2$), indicating conservative behavior suited for high-accuracy-critical applications, whereas other models undergo rapid set size reduction between $\alpha=0.1$ and $\alpha=0.2$ followed by stabilized decay rates, favoring efficiency-sensitive scenarios. Collectively, these results delineate $\alpha$'s systematic control over prediction set cardinality, while model-specific response characteristics inform context-optimized model selection.

\begin{figure}[!t]
\centering 

\begin{adjustbox}{width=\textwidth}
\begin{tabular}{ccc}
\subfloat[]{\includegraphics[width=5\textwidth]{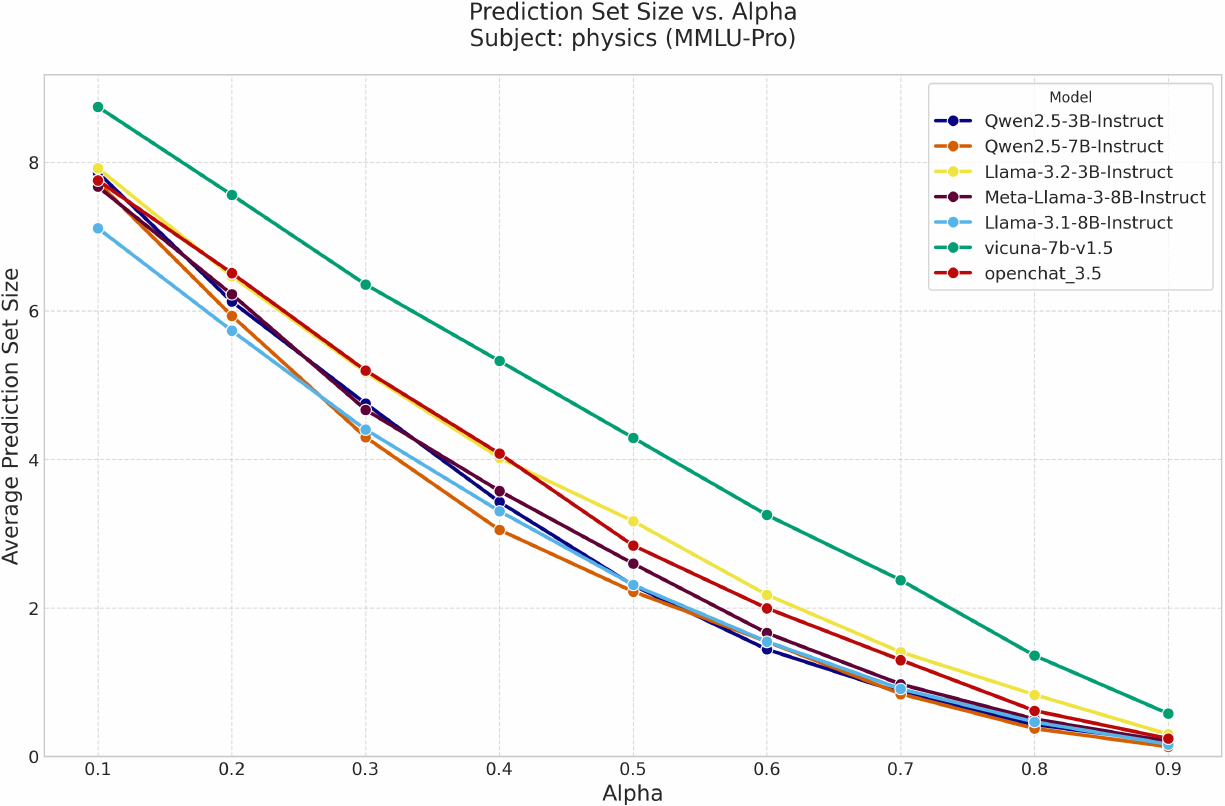}} &
\subfloat[]{\includegraphics[width=5\textwidth]{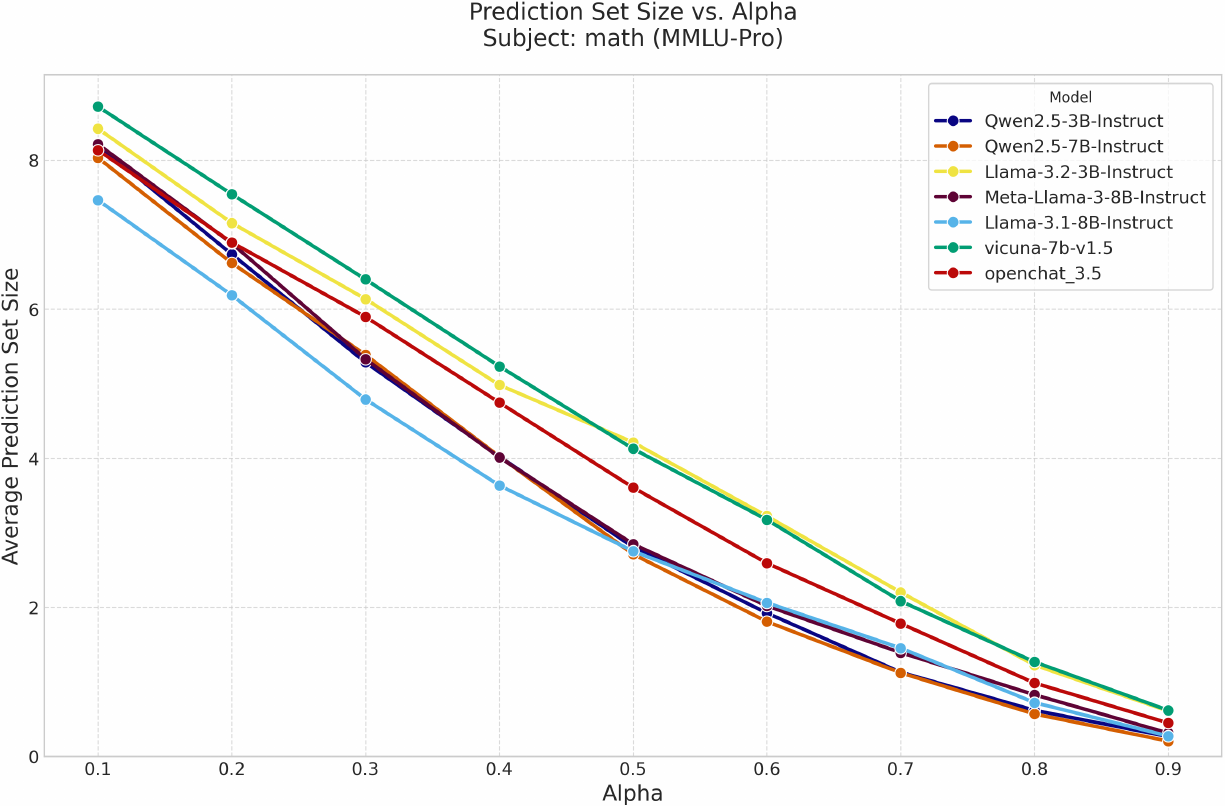}} \\
\subfloat[]{\includegraphics[width=5\textwidth]{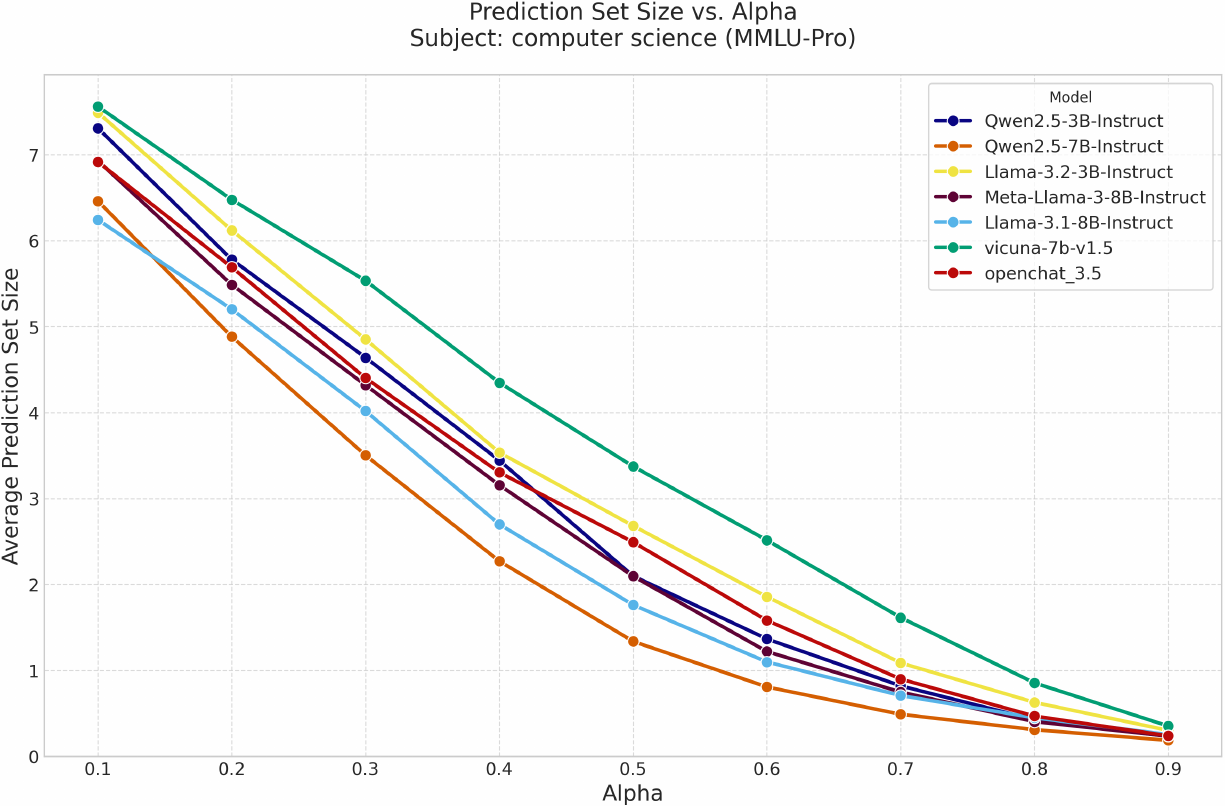}} &
\subfloat[]{\includegraphics[width=5\textwidth]{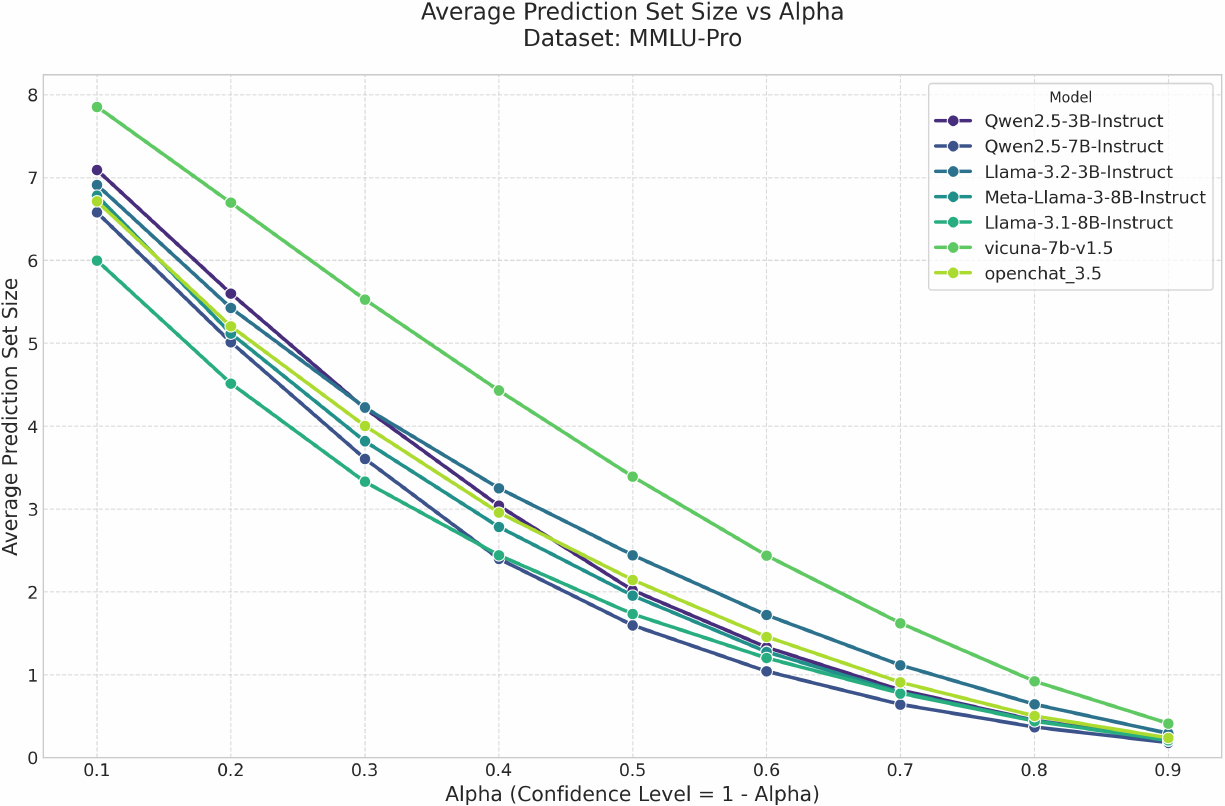}} \\
\end{tabular}
\end{adjustbox}

\caption{Comparative Curves of Individual Subject and Average Prediction Set Sizes in MMLU-pro}

\end{figure}

Similar to the MMLU dataset, MMLU-Pro exhibits decreasing prediction set sizes with increasing $\alpha$. However, the reduction is less pronounced between $\alpha=0.1$ and $\alpha=0.2$. Throughout $\alpha\in[0.1,0.5]$, set sizes decrease approximately linearly, with diminished reduction rates at higher $\alpha$ values. Analysis of individual models reveals Vicuna-7B-v1.5 maintains persistently largest prediction sets (indicating greater prediction variance), while other models demonstrate comparable performance with consistent cross-subject results, confirming the stability of our prediction set computation method.

Collectively, both datasets confirm that prediction set size contracts monotonically with increasing $\alpha$. Correlated with Section 4.2 results, this contraction implies elevated miscoverage rates. However, near $\alpha=0.4$, models converge to singleton prediction sets while maintaining low empirical error rates. This supports balancing miscoverage risk and set size at $\alpha\approx0.4$ for optimal selective prediction performance.

\section{Conclusion}

This study introduces an enhanced conformal prediction (CP) framework integrated with statistical significance testing to achieve risk control in domain-specific multiple-choice question answering (MCQA) tasks. The proposed framework provides statistically rigorous and interpretable methodology that mitigates model hallucination risks while enhancing the trustworthiness of large language models (LLMs) in disciplinary applications. Preserving CP's model-agnostic and distribution-free advantages, our approach simplifies computational procedures and extends functionality beyond basic coverage control to complex risk management scenarios. Crucially, the framework consistently achieves user-specified miscoverage rates ($\alpha$). Analysis of prediction set error rates across significance levels ($\alpha$) and corresponding set sizes reveals a well-defined inverse correlation between average prediction set size (APSS) and risk level, demonstrating APSS's utility for quantifying LLM uncertainty. Future work will investigate adaptation to other specialized QA domains and integration with advanced LLM architectures to further enhance reliability and transparency.

\bibliographystyle{plain}
\bibliography{references}

\appendix
\section{Proofs}
\begin{proof}
We establish the equivalence between the p-value formulation and conformal prediction (CP) through the following steps:

\textbf{Step 1: Conformal prediction set construction}\\
For a test point \(x_{N+1}\), the conformal prediction set at significance level \(\alpha\) is:
\begin{equation}
\mathscr{C}_{\alpha}(x_{N+1}) = \left\{ y : S(x_{N+1}, y) \leq Q_{1-\alpha}^{(N)} \right\}
\end{equation}
where \(Q_{1-\alpha}^{(N)} = \text{Quantile}\left( \{s_i\}_{i=1}^N, \frac{\lceil (N+1)(1-\alpha) \rceil}{N} \right)\) is the empirical quantile of conformity scores \(\{s_i = S(x_i, y_i^*)\}_{i=1}^N\) from the calibration set. This guarantees:
\begin{equation}
\mathbb{P}\left( y_{N+1}^* \in \mathscr{C}_{\alpha}(x_{N+1}) \right) \geq 1 - \alpha
\end{equation}
with probability over both calibration data \(\{(x_i, y_i)\}_{i=1}^N\) and test point \((x_{N+1}, y_{N+1}^*)\).

\textbf{Step 2: p-variable definition and equivalence}\\
Define the p-variable for candidate label \(y\):
\begin{equation}
P(y) = \frac{1}{N+1} \left( 1 + \sum_{i=1}^N \mathbb{I}\{s_i > S(x_{N+1}, y)\} \right)
\end{equation}
For the true label \(y = y_{N+1}^*\), we have:
\begin{equation}
P(y_{N+1}^*) = \frac{1}{N+1} \left( 1 + \sum_{i=1}^N \mathbb{I}\{s_i > s_{N+1}\} \right)
\end{equation}
where \(s_{N+1} = S(x_{N+1}, y_{N+1}^*)\). 

\textbf{Step 3: Coverage equivalence}\\
The inclusion condition \(y \in \mathscr{C}_{\alpha}(x_{N+1})\) is equivalent to:
\begin{align}
S(x_{N+1}, y) &\leq Q_{1-\alpha}^{(N)} \\
\iff \frac{1}{N} \sum_{i=1}^N \mathbb{I}\{s_i \leq S(x_{N+1}, y)\} &\leq \frac{\lceil (N+1)(1-\alpha) \rceil}{N} \\
\iff \sum_{i=1}^N \mathbb{I}\{s_i \leq S(x_{N+1}, y)\} &< \lceil (N+1)(1-\alpha) \rceil \\
\iff \sum_{i=1}^N \mathbb{I}\{s_i > S(x_{N+1}, y)\} &> N - \lceil (N+1)(1-\alpha) \rceil \\
\iff \frac{1}{N+1} \left( 1 + \sum_{i=1}^N \mathbb{I}\{s_i > S(x_{N+1}, y)\} \right) &> \frac{N - \lceil (N+1)(1-\alpha) \rceil + 1}{N+1} \\
\iff P(y) &> \alpha \quad \text{(for non-integer boundaries)}
\end{align}
Thus, \(y \in \mathscr{C}_{\alpha}(x_{N+1}) \iff P(y) > \alpha\).

\textbf{Step 4: Type I error control}\\
For the true label \(y_{N+1}^*\), we have:
\begin{align}
\mathbb{P}\left( P(y_{N+1}^*) \leq \alpha \right) &= \mathbb{P}\left( \frac{1}{N+1} \left( 1 + \sum_{i=1}^N \mathbb{I}\{s_i > s_{N+1}\} \right) \leq \alpha \right) \\
&= 1 - \mathbb{P}\left( \frac{1}{N} \sum_{i=1}^N \mathbb{I}\{s_i \leq s_{N+1}\} < \frac{\lceil (N+1)(1-\alpha) \rceil}{N} \right) \\
&\leq \alpha \quad \text{(by conformal coverage guarantee)}
\end{align}
This satisfies the p-variable condition \(\mathbb{P}(P \leq \alpha) \leq \alpha\) for all \(\alpha \in (0,1)\).

\textbf{Step 5: Hypothesis testing equivalence}\\
Consider the hypothesis test:
\begin{align*}
\mathscr{H}_0: & \quad y = y_{N+1}^* \quad \text{(true label)} \\
\text{Reject } \mathscr{H}_0 \text{ if} & \quad P(y) \leq \alpha
\end{align*}
The prediction set is exactly:
\begin{equation}
\mathscr{C}_{\alpha}(x_{N+1}) = \left\{ y : P(y) > \alpha \right\} = \left\{ y : \text{fail to reject } \mathscr{H}_0 \text{ for } y \right\}
\end{equation}
with miscoverage probability:
\begin{equation}
\mathbb{P}(y_{N+1}^* \notin \mathscr{C}_{\alpha}(x_{N+1})) = \mathbb{P}(P(y_{N+1}^*) \leq \alpha) \leq \alpha
\end{equation}

Therefore, the conformal prediction set construction is equivalent to a hypothesis testing framework using the defined p-variable, with identical coverage guarantees and error rate control.
\end{proof}

\end{document}